%% file: iclr2026_conference.tex
\documentclass{article} % For LaTeX2e
\usepackage{iclr2026_conference,times}

\input{math_commands.tex}

\usepackage{url}

\definecolor{iclrblue}{rgb}{0.21,0.49,0.74}
\usepackage[colorlinks, linkcolor=red, citecolor=iclrblue]{hyperref}

\usepackage{cleveref}

\usepackage{booktabs}       % professional-quality tables
\usepackage{amsfonts}       % blackboard math symbols
\usepackage{nicefrac}       % compact symbols for 1/2, etc.
\usepackage{microtype}      % microtypography
\usepackage{marvosym}       % \Letter symbol
\usepackage{xcolor}         % colors

\usepackage{makecell}
\usepackage{graphicx}
\usepackage{amsmath}
\usepackage{amssymb}
\usepackage{subcaption}  % 支持子图环境
\usepackage{diagbox}
\usepackage{multicol}
\usepackage{enumerate}
\usepackage{times}
\usepackage{epsfig}
\usepackage{threeparttable}
\usepackage{enumitem}
\usepackage{multirow}
\usepackage{color}
\usepackage{array}
\usepackage{setspace}
\usepackage{makecell}
\usepackage{subcaption}
\usepackage{csquotes}
\usepackage{adjustbox}
\usepackage{colortbl}
\usepackage{wrapfig}  
\usepackage{array}        % 用于列格式设置
\usepackage{graphicx}     % 用于 \resizebox
\usepackage{xcolor}       % 用于颜色设置
\usepackage{colortbl}     % 用于表格行颜色
\usepackage{multirow}     % 用于多行单元格
\usepackage{makecell}     % 用于 \makecell 命令
\usepackage{booktabs}     % 用于 \toprule 等命令
\usepackage{pifont} % 引入宏包
\definecolor{mygray}{gray}{0.6}
\definecolor{HTML}{RGB}{0.58,0,0}
\definecolor{myred}{RGB}{193, 39, 45}
\definecolor{tablered}{RGB}{193, 39, 45}
\definecolor{tableblue}{RGB}{0, 113, 188}
\definecolor{tablegreen}{RGB}{96, 181, 144}

\newcommand{\pub}[1]{\color{gray}{\tiny{[{#1}]}}}
\title{PreferThinker: Reasoning-based Personalized Image Preference Assessment}

\author{
  \begin{tabular}{c} % 使用 tabular 环境并指定 c (居中)
    Shengqi Xu$^{\spadesuit}$, 
    Xinpeng Zhou$^{\clubsuit}$, 
    Yabo Zhang$^{\clubsuit}$, 
    Ming Liu$^{\clubsuit,}$\textsuperscript{\Letter}, 
    Tao Liang$^{\blacklozenge}$,
    Tianyu Zhang$^{\blacklozenge}$
    \\ \vspace{3mm} % 注意：\And 不能在 tabular 里工作，用 \\ 替换
    Yalong Bai$^{\blacklozenge}$,
    Zuxuan Wu$^{\spadesuit}$,
    Wangmeng Zuo$^{\clubsuit}$
    \\
    \vspace{3mm}
    $^{\spadesuit}$ Fudan University \hspace{0.3cm}
    $^{\clubsuit}$ Harbin Institute of Technology \hspace{0.3cm}
    $^{\blacklozenge}$ iN2X
    \\
    \href{https://3038543815.github.io/preferthinker.github.io/}{Project Page}
  \end{tabular}
}

\iclrfinalcopy % Uncomment for camera-ready version, but NOT for submission.
\begin{document}

\maketitle

\input{Secs/abstract}
\input{Secs/intro}

\input{Secs/related_work}

\input{Secs/method_dataset}

\input{Secs/experiments}

%================================introduction==========================================

\vspace{-5pt}
\section{Conclusion}
\vspace{-5pt}
 We introduce PreferThinker, the first  reasoning-based personalized image preference assessment system with preference profile prediction. We propose a preference profile to bridge various uses,  allowing large-scale
user data to be leveraged for training profile prediction and capturing complex
personalized preferences. We construct PreferImg-CoT, a CoT-style  dataset annotated with  preference profiles and high-quality reasoning for interpretability supervision. We adopt a two-stage training strategy comparing  Cold-start SFT  and reinforcement learning to empower the model with reasoning capabilities.  We propose a similarity-aware prediction reward to improve profile prediction, facilitating reasonable assessments. Extensive experiments verify the superiority of PreferThinker.

\section*{Acknowledgments}
This work was supported by National Key RD Program of China under Grant No. 2022YFA1004100.

\bibliography{iclr2026_conference}
\bibliographystyle{iclr2026_conference}

\appendix

\input{Secs/appendix}

\end{document}

%% file: math_commands.tex
\usepackage{amsmath,amsfonts,bm}

\def\eqref#1{equation~\ref{#1}}
\def\1{\bm{1}}

\DeclareMathAlphabet{\mathsfit}{\encodingdefault}{\sfdefault}{m}{sl}
\SetMathAlphabet{\mathsfit}{bold}{\encodingdefault}{\sfdefault}{bx}{n}

%% file: Secs/abstract.tex
\vspace{-7mm}
\begin{abstract}
% 个性化偏好评估任务是一个很难的任务，只能有少量的个性化图像来辅助评估待选图像对于每一个用户
%  大多数现有的方法通常使用大量的通用人类偏好数据来微调CLIP或者MLLM来评估已经定义好的通用人类偏好（e.g. text-image alignment）。然而这类方法通常无法处理个性化偏好 由于个性化数据之间具有差异性导致的数据有限以及个性化偏好的复杂
%针对以上问题，我们引入了一个公共的偏好准则作为桥梁来减少用户之间的gap从而可以使用大量用户样本进行训练，并刻画复杂的个性化偏好
%此外，大多数现有的方法仅输出一个数值分数，缺乏可解释性
%在这个工作中，我们提出了一个基于推理的个性化评估系统并带有多维度偏好准则预测。具体来说，我们首先基于个性化参考图像来预测出用户的偏好准则，然后再基于预测的准则来对待选图像进行可解释性和多维度的评估。
%To this end，我们首先构建了一个大规模的思维链风格-个性化评估数据集，标注了丰富的多维度偏好准则和Cot-style的推理， enabling explicit supervision of structured reasoning.
%Next，We introduce a training strategy comprising cold-start and GRPO-based RFT to enables the systemto master the reasoning capability. Furthermore, we propose a similarity-aware prediction rewardto better predict the user’s preference criteria, facilitating more reasonable evaluation exploration.
%在多个数据集的实验验证了我们的方法的有效性

Personalized image preference assessment aims to evaluate an individual user's image preferences  by relying only on a small set of reference images as prior information. Existing methods mainly focus on general preference assessment, training models with large-scale data to tackle well-defined tasks such as text-image alignment and aesthetics. However, these approaches struggle to handle personalized preference assessment because user-specific data are typically scarce and not easily scalable, and individual tastes are often diverse and complex. To overcome these challenges, we introduce a common preference profile that serves as a bridge across  users, allowing large-scale user data to be leveraged for training profile prediction and  capturing complex personalized preferences.  Building on this idea, we propose a reasoning-based personalized image preference assessment framework that follows a \textit{predict-then-assess} paradigm:  it first predicts a user's preference profile from reference images, and then provides interpretable, multi-dimensional scores and assessments of candidate images based on the predicted preference profile. To support this, we first construct a large-scale Chain-of-Thought (CoT)-style personalized assessment dataset annotated with diverse user preference profiles and high-quality CoT-style reasoning, enabling explicit supervision of structured reasoning. Next, we adopt a two-stage training strategy: a cold-start supervised fine-tuning  phase to empower the model with  structured reasoning capabilities, followed by reinforcement learning to incentivize the model to explore more reasonable assessment paths and enhance generalization. Furthermore, we propose a similarity-aware prediction reward to encourage better prediction of the user's preference profile, which facilitates more reasonable assessments exploration. Extensive experiments demonstrate the superiority of the proposed method.
\end{abstract}

%% file: Secs/intro.tex
  \vspace{-5mm}
\section{Introduction}

%================================任务背景和难点==========================================
%1. 图像偏好评估对于评估和对齐生成模型与人类对齐很重要，在生成内容推荐
%2  现有的方法主要是聚焦于已经定义好的通用人类偏好评估，如文本-图像对齐和美学等，并且可以依赖于容易大量扩展的通用人类偏好数据。
%3. 然而忽略了具有挑战并且实用的个性化偏好评估的任务，个性化评估的任务设定
%4. 个性化评估相比于通用偏好评估的难点

\begin{figure}[t]
    \vspace{-3mm}
  \setlength{\abovecaptionskip}{0pt}
  \setlength{\belowcaptionskip}{-15pt}
  \captionsetup{skip=2pt} % 仅对当前图片生效
  \centering
   \includegraphics[width=1\linewidth]{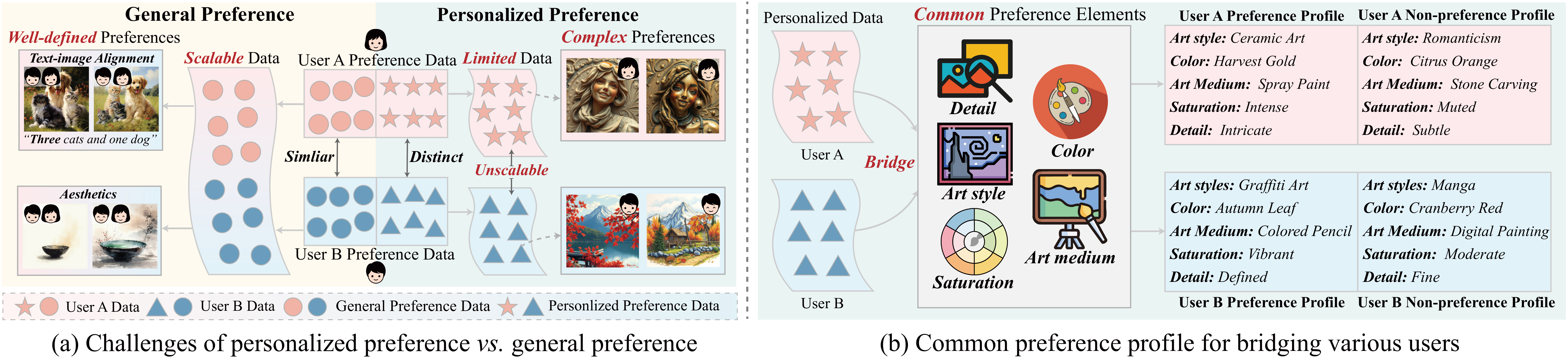}
   \caption{Illustration of challenges and motivation. (a) The general preference data is easily scalable since users share common assessment criteria, while personalized preference data for each user is typically limited and unscalable, as each user's preferences are distinct.  Besides, general preferences  are  often clear (\textit{e.g.} text-image alignment and aesthetics), while personalized preferences are typically complex and diverse.  (b) We propose a preference profile comprising multiple common visual elements, based on the observation that  although each user's personalized preferences are unique, the key visual elements that shape them are shared and can therefore serve as a bridge to connect users.}
   \label{diff}
  % \vspace{-1mm}
\end{figure}

Image preference assessment is essential for both evaluating and aligning generative models with human preferences, playing a key role in applications like content recommendation. However,  most existing works \citep{kirstain2023pick,xu2023imagereward} focus on assessing \textit{general} preferences, such as text-image alignment and aesthetics, relying on large-scale general preference data. The practical but challenging task of \textit{personalized} preference assessment remains largely underexplored, where only a limited amount of personalized data is available for each user as prior information to assist in the assessment. As shown in Fig. \ref{diff} (a), personalized preference assessment is particularly challenging for two main reasons. First, unlike \textit{easily scalable} general preference data, where most users share similar assessment criteria, each user's personalized data is typically \textit{limited} and \textit{distinct} from others, making large-scale, user-specific training difficult.  Second, personalized preferences are often  \textit{complex} and \textit{diverse}, covering multiple dimensions such as art style, color, and art medium, which are more challenging than \textit{clear} general human preferences (\textit{e.g.}  text-image alignment and aesthetics).

%TODO: 图1需要写caption 以及 需要加一个思路图 放在右边  展示我们得方法核心  比如引入一个判断标准 去桥接不同的用户 
%================================现有工作缺陷和问题==========================================
%1. 现有方法主要是分为两类
%2. CLIP-based方法，fine-tune CLIP模型，评估通用人类偏好，缺点是难以处理个性化评估，且缺乏可解释性
%3. MLLM-based方法，fine-tune MLLM模型，评估
%4. 现有的一个方法ViPer，直接将个性化参考图像和候选图像输入到MLLM中进行评估，缺点是隐式利用参考图像，缺乏可解释性                                                   
Current research on image preference assessment mainly falls into two categories: \textit{CLIP-based} methods and \textit{Multimodal Large Language Model (MLLM)-based} methods. 
\textit{CLIP-based} methods \citep{wu2023human,wu2023human2} typically fine-tune the CLIP model using large-scale text-image pairs with similarity to assess general preferences. 
However, they struggle with personalized preferences since the amount of personalized data  is typically limited for large-scale training.  Besides, simple similarity  fails to capture complex individual preferences and lacks interpretability,  as it  outputs only a numerical score. 

\textit{MLLM-based} methods \citep{wang2025unified,xu2024visionreward} achieve interpretable preference assessment by fine-tuning the MLLM on large-scale visual question-answering (VQA) pairs. However, the scarcity of personalized images makes it difficult to obtain sufficient VQA pairs for large-scale training, which limits their effectiveness for personalized preference assessment. Overall, most existing methods are designed for general preferences and therefore struggle to handle personalized preference. Recently, an MLLM-based method named ViPer \citep{salehi2024viper} was proposed for personalized preference assessments. It directly feeds both personalized reference images and candidate images into an MLLM and  fine-tunes the model using a supervised score regression strategy. Nevertheless, it only \textit{implicitly} utilizes reference images to assist in evaluating the candidate images, failing to fully leverage the critical prior information they contain. Moreover, it lacks interpretable reasoning steps that explain how to produce its final score based on the individual reference images.

%%================================我们的工作和方法==========================================

% 为了解决以上的难题，在这个工作中，我们提出了一个基于推理的个性化偏好评估方法，并且带有属性级别的偏好标准预测。
% 个性化偏好
%到个性化偏好评估的挑战性和复杂性，我们提出了一个基于推理的个性化偏好评估方法，PreEval。该方法结合了显式的视觉偏好预测和多维度的链式思维（CoT）推理。具体而言，PreEval首先从个性化参考图像中预测用户的视觉偏好，然后基于预测的视觉偏好对候选图像进行多维度评估。与现有方法相比，PreEval不仅可以实现更准确的评估，还能提供基于用户个性化偏好的可解释数值评分。
% 为解决上述挑战，我们提出了一个基于推理的个性化评估系统，其核心在于显式的偏好准则预测和多维度的思维链式评估。

% 我们的核心思想是引入一个由五个关键视觉属性构成的跨用户评估准则。虽然每个用户的最终偏好是独特的，但构成这些偏好的基础属性（如色彩、构图、艺术风格等）是通用的，可以作为桥梁连接不同用户。这个方法一方面通过预测用户对这些属性的偏好来刻画其个性化视觉偏好，从而解决了偏好难以定义的难题。另一方面，由于这些基础属性可以跨用户共享和学习，我们能够利用大规模用户样本进行训练，以有效地预测每个用户的独特评估准则，从而缩小了用户数据之间的鸿沟。

% 具体来说，我们首先利用用户的个性化参考图像，预测其独特的视觉偏好标准。随后，我们使用这些标准去评估待选图像，并给出可解释性的得分。

%每一个人都有以一个自己的评判标准
%我们为每个用户都制定一个偏好评判标准来bridge每一个用户  这个评判标准可以跨用户使用 视觉属性级别的

To address the above issues, we propose a reasoning-based personalized preference assessment system  with preference profile prediction, named \textbf{PreferThinker}. Our key insight is to bridge various users by introducing a preference profile comprising multiple visual elements (\textit{e.g.} color, art style), as shown in Fig \ref{diff}(b). \textit{Although each user's personalized preferences are unique, the fundamental elements that form them are common and can serve as a bridge across users}. This profile offers three main advantages. First,  the element-level profile allows us to characterize personalized preferences, which alleviates the challenge of complex individual tastes. Second, since these visual elements can be shared and learned across  users, we can leverage large-scale user samples for training to effectively predict each user's  visual preference profile, thereby mitigating the issue of limited personalized data. Third, it provides a solid foundation for subsequent interpretable and multi-dimensional assessments.
%可以通过这些判别标准去进行可解释性地打分

\begin{figure}[t]
  \vspace{-7.4mm}
  \setlength{\abovecaptionskip}{0pt}
  \setlength{\belowcaptionskip}{-15pt}
  \captionsetup{skip=2pt} % 仅对当前图片生效
  \centering
   \includegraphics[width=1\linewidth]{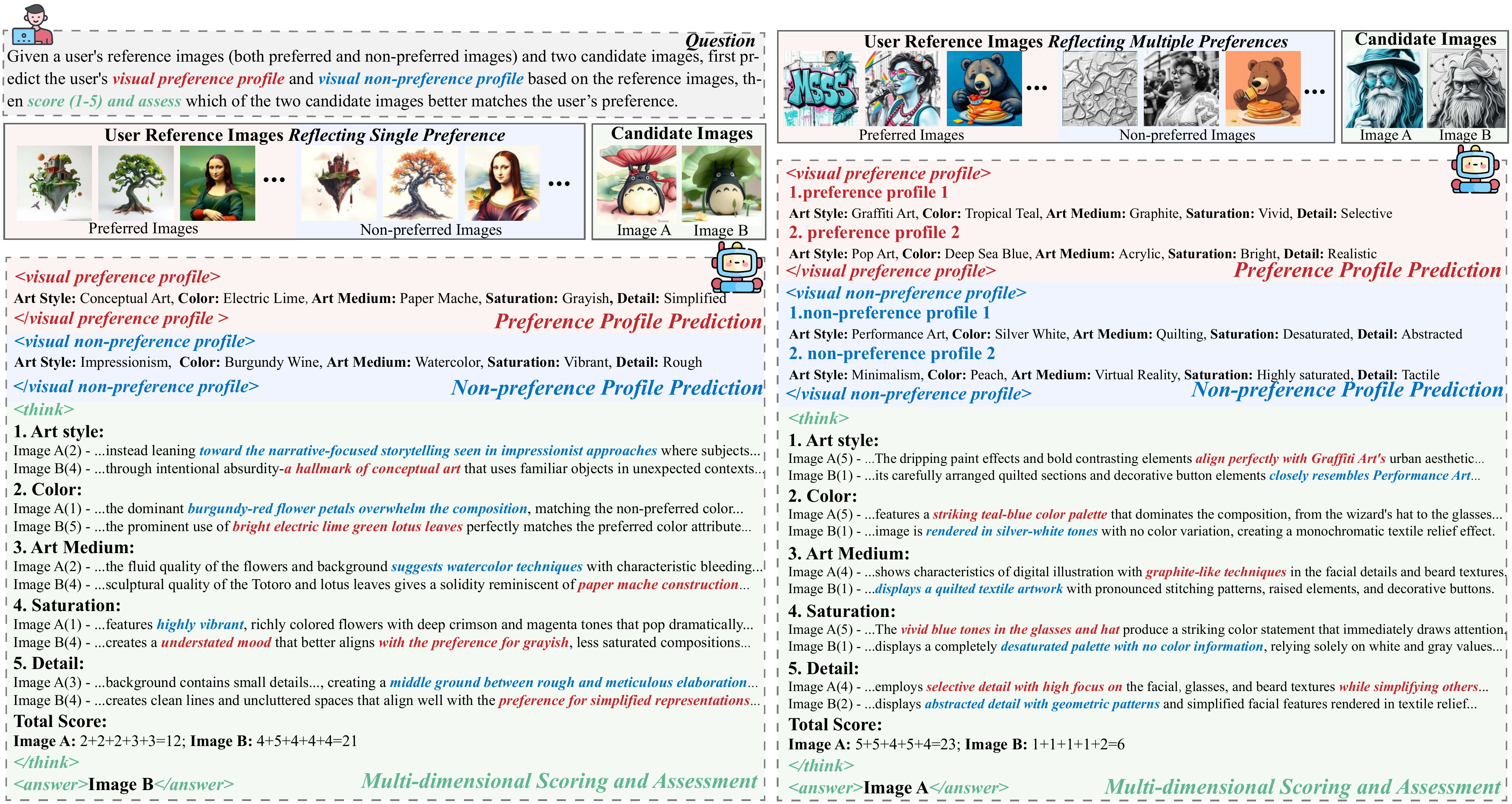}
   \caption{Examples of PreferThinker for personalized image preference assessment.  In the \textbf{\textit{\textcolor{tablegreen}{think}}} stage, \textbf{\textit{\textcolor{tablered}{red text}}} denotes alignment with preference profiles, while \textbf{\textit{\textcolor{tableblue}{blue text}}} denotes alignment with non-preference profiles. \textit{See Appendix  \ref{sec:supp_visualization} for more complete reasoning examples.} }
   \label{demo}

\end{figure}

Specifically,  given a user's personalized reference images (preferred and non-preferred) and two candidate images, where reference images may reflect multiple aspects of user's tastes. PreferThinker follows a  \textit{predict-then-assess} CoT-style structure for personalized preference assessment. As shown in Fig. \ref{demo}, the process consists of two stages:  1) Profile Prediction: PreferThinker first predicts the user's visual preference and non-preference profiles comprising multiple visual elements based on the reference images. 2) Multi-dimensional and Interpretable Assessment: Using the predicted profile, PreferThinker provides interpretable scores for the candidate images across multiple dimensions and produces a final result based on the total score. This \textit{predict-then-assess} framework not only enables accurate assessment, but also achieves interpretable scoring grounded in the user's reference images.

% \begin{figure}[t]
%     \vspace{-7.5mm}
%   \setlength{\abovecaptionskip}{0pt}
%   \setlength{\belowcaptionskip}{-15pt}
%   \captionsetup{skip=2pt} % 仅对当前图片生效
%   \centering
%    \includegraphics[width=1\linewidth]{Figures_pdf/Figure1_3.pdf}
%    \caption{Illustration of challenges and motivation. (a) The general preference data is easily scalable since users share common assessment criteria, while personalized preference data for each user is typically limited and unscalable, as each user's preferences are distinct.  Besides, general preferences  are  often clear (\textit{e.g.} text-image alignment and aesthetics), while personalized preferences are typically complex and diverse.  (b) We propose a preference profile comprising multiple common visual elements, based on the observation that  although each user's personalized preferences are unique, the key visual elements that shape them are shared and can therefore serve as a bridge to connect users.}
%    \label{diff}
% \end{figure}

% This allows us to use large-scale user samples for training and to approximately characterize personalized preferences, while the profile facilitate interpretable and multi-dimensional assessments.

%============================数据集，训练策略，奖励设计=============================
%1. 为了让模型掌握先预测再评估的思维链方式的架构
%2. 我们首先构建了一个大规模的思维链式个性化评估数据集，该数据集包含了XXX个用户样本，和多少张图像，每一个用户样本都带有多维度的视觉偏好标准标注以及思维链风格的评估标注，有利于后续对step-by-step reasoning的显式监督。
%3. 然后我们引入了一个两阶段的训练策略，首先是冷启动的监督微调阶段，让模型学习如何进行结构化推理，然后是基于GRPO的强化微调阶段，引导模型探索更合理的评估路径，提升泛化能力。
%4. 此外，我们发现更好的偏好标准预测也有助于探索更可解释的评估过程，这促使我们提出了一个相似度感知的预测奖励函数，以在强化学习阶段更好地预测用户的视觉偏好。
To this end, we first construct a large-scale CoT-style personalized image preference assessment dataset, named \textbf{PreferImg-CoT}, which contains 60,000 user samples, annotated with preference profiles and high-quality CoT-style assessments, enabling explicit supervision of structured reasoning. Next, we adopt a two-stage training strategy: a cold-start supervised fine-tuning (SFT) phase to elicit the model structured reasoning, followed by reinforcement fine-tuning (RFT) based on Group Relative Policy Optimization (GRPO) to guide model toward exploring more reasonable assessments and enhancing  generalization.
Moreover, we find that accurate preference profile prediction facilitates exploring reasonable assessments, which motivated us to propose a  similarity-aware prediction reward to better predict user's preference profile. Overall, our main contributions are summarized as follows:

  \vspace{-3mm}
\begin{itemize}[leftmargin=10pt]
    \item  We propose PreferThinker, a reasoning-based personalized image assessment system with preference profile prediction, which achieves accurate assessment along with interpretable and multi-dimensional scoring based on the predicted profile. 
    \setlength{\itemsep}{0pt}
    \item We construct PreferImg-CoT, the first CoT-style personalized assessment dataset, annotating with preference profile and CoT-style reasoning, facilitating the supervision of model interpretability.
    \setlength{\itemsep}{0pt}
    \item We adopt a training strategy comprising cold-start and GRPO-based RFT to enables the system to master the reasoning capability. Furthermore, we propose a similarity-aware prediction reward to better predict the user's  preference profile, facilitating more reasonable assessments exploration.
    \item We  compare PreferThinker with existing methods on the proposed and existing datasets. Extensive experiments confirm that PreferThinker outperforms SOTA methods. Moreover, the predicted preference profile of PreferThinker can also benefit personalized image generation.

\end{itemize}
%================================introduction=================================

% \begin{figure}[t]
%     \vspace{-7.4mm}
%   \setlength{\abovecaptionskip}{0pt}
%   \setlength{\belowcaptionskip}{-15pt}
%   \captionsetup{skip=2pt} % 仅对当前图片生效
%   \centering
%    \includegraphics[width=1\linewidth]{Figures_pdf/Figure2_2.pdf}
%    \caption{Examples of PreferThinker for personalized image preference assessment.  In the \textbf{\textit{\textcolor{tablegreen}{think}}} stage, \textbf{\textit{\textcolor{tablered}{red text}}} denotes alignment with preference profiles, while \textbf{\textit{\textcolor{tableblue}{blue text}}} denotes alignment with non-preference profiles. \textit{See Appendix  \ref{sec:supp_visualization} for more complete reasoning examples.} }
%    \label{demo}
% \end{figure}

%% file: Secs/related_work.tex
  \vspace{-5mm}
\section{Related Work}
\label{gen_inst}

%================================图像偏好评估==========================================
%1. 图像偏好评估的意义
%2. 传统的评估方法如FID和CLIP score等，无法捕捉人类偏好
%3. 为了解决这个问题，一些工作通过使用大规模的人类通用偏好数据来微调CLIP或者BLIP模型,从而来评估通用偏好，如文本一致性。然而这些方法缺乏可解释性由于只输出一个数值分数。为了解决这个问题，一些基于MLLM的方法应运而生，主要受益于MLLM强大的多模态理解能力，如xu和wang使用大量的视觉问答对来微调MLLM以实现可解释性地评估。
%4.然而大多数的方法都是聚焦于通用偏好评估，难以处理个性化偏好评估，因为用户个性化数据的有限以及个性化偏好的复杂性。 Viper stand out 提出了一个基于MLLM的方法来进行个性化评估，然而该方法简单地将参考图像和候选图像输入到MLLM中进行监督的数值分数回归，缺乏可解释性。
%5.在这个工作中，我们提出了一个基于推理的个性化评估系统并且带有显式地视觉偏好标准预测，我们引入了多维度偏好准则来桥接不同的用户，我们既能够实现精确的个性化偏好评估，又能够实现基于偏好标准的可解性打分。
\textbf{Image Preference Assessment} is crucial for both evaluating and aligning generative models with human preferences. While metrics such as FID \citep{heusel2017gans} and CLIP scores \citep{radford2021learning} assess image quality and text-image consistency, they are limited in capturing human preferences. To address this, most works \citep{kirstain2023pick,liang2024rich,li2024stable,zhang2024learning,ba2025enhancing, chen2025pal} employ  large-scale preference data to fine-tune CLIP models for assessing general preferences. However, they lack interpretability, acting as black boxes that  output only a score. Recently, MLLM-based methods\citep{zhou2025multimodal, wang2025unifiedthink,mo2025learning,gambashidze2025listener} have emerged, benefiting from the powerful multimodal understanding of MLLMs. For example, \citep{xu2024visionreward} and \citep{wang2025unifiedthink} fine-tune MLLMs using large-scale visual question-answering pairs to achieve interpretable assessments. Nevertheless, most existing methods focus on general preferences and struggle with personalized preferences due to the scarcity of personalized data and complexity of individual tastes. Though \citep{salehi2024viper} stands out by proposing a personalized assessment method, it naively feeds reference and candidate images into an MLLM for score regression, thus lacking interpretability. In this work, we propose a reasoning-based personalized assessment system that achieves both accurate assessments and interpretable scoring.

\textbf{Image Preference Assessment Dataset} is essential for training preference assessment models. \citep{xu2023imagereward} proposes ImageRewardDB, the first dataset for training models to assess general preferences (\textit{e.g.} text-image alignment, aesthetics and safety), containing 137K candidate image pairs. Further, \citep{kirstain2023pick} and \citep{wu2023human2} respectively developed PickaPic\_v2 and HPD\_v2, two large-scale and diverse datasets for learning human preferences. However, most existing public datasets focus on training models to capture general preferences, with limited consideration for personalized preferences. While \citep{salehi2024viper} curates a simulated personalized preference dataset, it remains unreleased. In this work, we construct a large-scale personalized assessment dataset annotated with diverse preference profiles and CoT-style reasoning to advance this field.

\textbf{Reasoning-based Multimodal Large Language Models} have achieved significant progress in multimodal reasoning abilities through CoT-based or RL-based fune-tuning.  Deepseek-R1 \citep{guo2025deepseek} employs the GRPO algorithm \citep{shao2024deepseekmath} with the rule-based reward to improve reasoning abilities of LLM without critic models. Inspired by this, several works have incorporated GRPO-based post-training into MLLMs to enhance the multimodal reasoning in various tasks, including  visual perception \citep{jiang2025rex,huang2025vision,liu2025visual,zhang2025improving,yu2025perception,liu2025seg,you2025seg} and visual understanding \citep{chen2025chart,pan2025medvlm,ni2025point}, math problem solving \citep{guo2025decoupled,zhang2025r1,li2025vision},  and visual quality assessment \citep{wu2025visualquality,li2025q,cai2025q}.
Closest to ours, \citep{wang2025unifiedthink} explores reasoning-based image preference assessment through RL-based post-training. However, this method focuses on assessing general preferences. Distinctly, PreferThinker is the first reasoning-based personalized assessment system. Inspired by Deepseek-R1,  we adopt a two-stage training strategy comprising cold-start SFT and GRPO-based RFT to enable the system to master the reasoning capability Furthermore, we propose a similarity-aware prediction reward to better predict user's preference profile, facilitating  more reasonable evaluations exploration.

%% file: Secs/method_dataset.tex
  \vspace{-3mm}
\section{Reasoning-based Personalized Image Preference Assessment}
\label{others}

%=====================Overview of the method========================

%=====================Overview of the method========================
In this work, we propose PreferThinker, a reasoning-based  assessment system  with preference profile prediction.  Given a few  individual reference images (both preferred and non-preferred) and two candidate images, PreferThinker assesses the candidate images following a \textit{predict-then-assess} CoT-style structure. As shown in Fig. \ref{demo}, it first predicts the user's visual preference and non-preference profiles based on the reference images. Next, it utilizes the predicted profiles as criteria to  score the candidate images across multiple dimensions in an interpretable manner and produces a final result based on the total score. By leveraging this \textit{predict-then-assess} paradigm, PreferThinker  not only achieves accurate assessment but also providing interpretable scoring based on the predicted profiles.

Specifically, we first introduce a preference profile composed of multiple common visual elements to bridge various users (Section \ref{sec3.1}). Next, we construct a large-scale CoT-style  dataset annotated with preference profiles and high-quality reasoning to provide reasoning supervision (Section \ref{sec3.2}). We  then employ a two-stage training strategy  comprising cold-start SFT and RL-based post-training to enable the system
to master the reasoning capability (Section \ref{sec3.3}).  Moreover, we discover that more accurate
 profile prediction facilitates exploring more reasonable assessments, which motivates us to propose a similarity-aware prediction reward to better predict preference profile (Section \ref{sec3.4}).
%=====================Overview of the method========================

\vspace{-5pt}
\subsection{Visual Preference Profile}
\label{sec3.1}
%common preference criteria动机是什么
%大部分现有的偏好评估方法主要使用大规模和容易扩展的通用偏好数据来训练模型实现通用偏好评估。然而他们无法处理个性化偏好评估方法。因为每个用户的个性化偏好数据是有限的，并且用户之间的差异导致其难以扩展。同时个性化偏好是难以定义的。为了解决以上问题，我们提出了一个通用的偏好标准来缩小用户之间的gap。该标准由多个关键视觉属性组成，这些属性在很大程度上影响用户对图像的偏好。我们argue that尽管每个用户的个性化偏好是独特的，但是形成这些偏好的关键视觉属性是通用的并且阔以作为桥梁连接不同用户的个性化偏好。

Most existing methods mainly rely on large-scale, easily scalable general preference data to train models for general preference assessment. However, they struggle to handle personalized preference due to the limited personalized data for each user and user heterogeneity. To address  this issue, we propose a preference profile comprising multiple common visual elements to bridge various users. We argue that \textit{although each user's personalized preferences are unique, the key visual elements that shape these preferences are shared and can serve as a intermediate bridge to connect various users}. 
%common preference criteria是怎么定义的
%由于个性化偏好是抽象和复杂的，为了简化，我们首先识别出15个在Lexica的prompts中经常出现并且最能影响用户对生成图像偏好的视觉属性。然后我们进行了一项调查，邀请100名参与者，每人选择五个最重要的视觉属性作为偏好准则。结果显示，艺术风格、颜色、细节、艺术媒介和饱和度被选为偏好准则的最具代表性的视觉属性。同时为了确保个性化的多样性，我们为每个属性收集了丰富的相关词汇，为后续构建大规模和多样的个性化数据集打下了基础，如图\ref{person_data}(b)所示。

To formalize a user's complex personalized visual preferences, we  identify 15 visual elements that most frequently appear in Lexica's \footnote{\href{https://lexica.art/}{https://lexica.art/}} text prompts and strongly influence user preference toward personalized image generation. We then conduct a user study with 100 participants, asking each to select the five most important visual elements as visual preference profile. The result (See Fig \ref{supp_criteria} in the Appendix) reveals that \textit{art style}, \textit{color}, \textit{detail}, \textit{art medium} and \textit{saturation} are voted as the most representative visual elements for characterizing personalized preferences. To ensure preference profile diversity, we collect a rich vocabulary of related terms for each visual element, totaling 288 words (See Fig. \ref{supp_vocabulary} in the Appendix), laying a solid foundation for constructing a large-scale and diverse dataset for personalized image preference assessment.
%common preference criteria的优势是什么
%所提出的通用偏好准则有以下优势, This criteria provides three main advantages. First, we
% utilize attribute-level criteria to characterize personalized preferences, which alleviates the challenge
% of ill-defined personalized preferences. Second, since these elements can be shared and learned
% across users, we can leverage large-scale user samples for training to effectively predict each user’s
% assessment criteria, thereby mitigating the issue of limited personalized data. Third, it provides a
% solid foundation for subsequent interpretable, multi-dimensional assessment.

\begin{figure}[t]
  \vspace{-6mm}
  \setlength{\abovecaptionskip}{0pt}
  \setlength{\belowcaptionskip}{-15pt}
    \captionsetup{skip=2pt} % 仅对当前图片生效 
  \centering
   \includegraphics[width=1\linewidth]{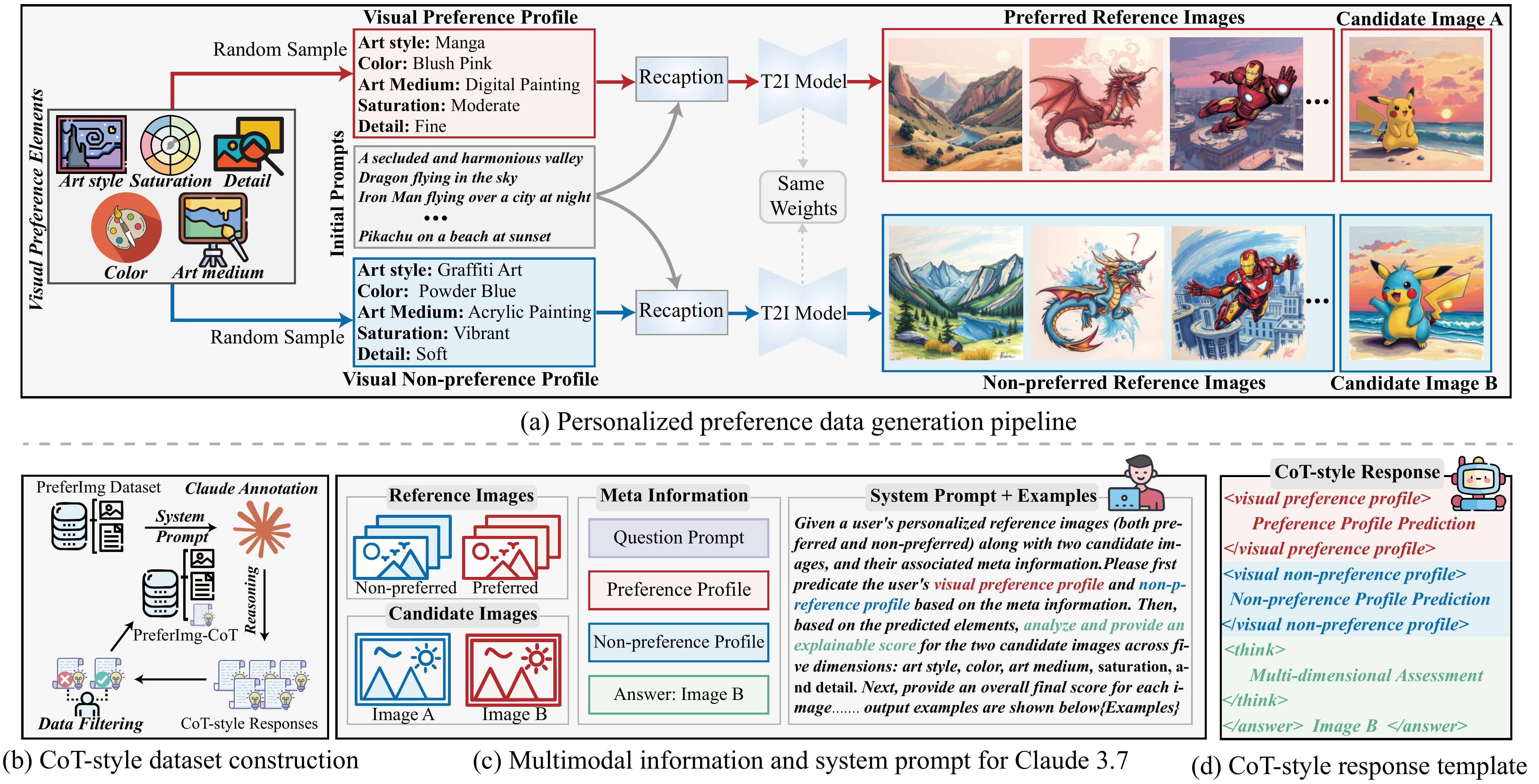}
   \caption{Illustration of the proposed dataset PreferImg-CoT. (a) Personalized preference data generation pipeline. (b) Overview of CoT-style dataset construction: Claude annotation and data filtering. (c) Prompt design for Claude 3.7 to generate CoT-style response. (d) CoT-style response template, including preference profiles prediction, multi-dimensional assessment and answer.
   }
   \label{personalized_cold_data}
     \vspace{-1mm}
\end{figure}

\subsection{Large-scale CoT-style Personalized Preference Assessment Dataset}
\label{sec3.2}
  \vspace{-2mm}

To enable the model to master CoT-style reasoning capabilities, we construct a large-scale CoT-style dataset that provides high-quality reasoning supervision. Since most existing public datasets focus on \textit{general} preference assessment, we first construct a new dataset called PreferImg, dedicated to \textit{personalized} image preference assessment, annotated with diverse visual preference profiles. We then employ the advanced  Claude 3.7 model \citep{anthropic2024claude} to generate high-quality CoT-style annotations on the proposed PreferImg, ultimately producing our  CoT-style dataset, PreferImg-CoT. 
%构建COT-style的动机是什么？
%动机：为了让个性化评估模型能够拥有先预测视觉偏好再评估的能力
%怎么构建这个数据集的？
%数据集特性是什么样子有什么优势  大规模  高质量  带有个性化评估

%视觉偏好属性

\textbf{Personalized  Preference Assessment Data Generation.} Due to the difficulty in acquiring
personalized preference data with annotated preference profiles, such datasets are currently unavailable.  To fill this gap, inspired by \citep{mo2025prefgen}, we construct a large-scale personalized assessment dataset containing 80K simulated users, each with a distinct preference profile. The data generation pipeline is shown in the Fig. \ref{personalized_cold_data}(a). Specifically,  we  first randomly sample five  visual preference elements to assign visual preference and non-preference profiles to each user. To account for the fact that real users may have multiple personalized preferences, several preference profiles are assigned to a subset of users. Then, we combine these profiles with initial prompts and feed into a \text{text-to-image} model to generate each user's reference images (preferred and non-preferred) and two candidate images.  To further guarantee content diversity, we follow the PrefGen \citep{mo2025prefgen}to select 190 K prompts from Lexica, DiffusionDB \citep{wang2022diffusiondb}, and COCO \citep{lin2014microsoft} as initial prompts. Ultimately, we establish a dataset annotated with preference profiles, comprising 80 K users, 20K of whom are associated with multiple preference profiles, along with 1.36 million images, laying a solid foundation for subsequent CoT-style dataset construction.

%刻画用户的个性化视觉偏好是一个复杂的问题。为了在保证准确性的同时降低问题复杂度，我们基于用户调研和领域知识，遴选了五个最具代表性的视觉属性作为个性化偏好描述的核心维度

%由于获取带有精细视觉属性标注的真实用户个性化偏好数据存在较大困难，据我们所知，目前尚未有相关公开数据集。尽管ViPer构建了一个仿真用户个性化评估数据集，但该数据集并未公开。为此，本研究参考ViPer，构建了包含XXX个仿真用户的偏好评估数据集。这些仿真用户的个性化特征通过其偏好与非偏好的视觉属性进行刻画。刻画用户的个性化视觉偏好是一个复杂的问题。为了在保证准确性的同时降低问题复杂度，我们基于用户调研和领域知识，遴选了五个最具代表性的视觉属性作为个性化偏好描述的核心维度

%

\textbf{CoT-style Dataset Construction.} To construct a  high-quality CoT-style dataset, we design a two-stage  construction process: Claude annotation and data filtering, as shown in Fig. \ref{personalized_cold_data}(b). First, we employ  Claude 3.7 with prompting to generate CoT style responses. Figure \ref{personalized_cold_data}(c) shows that the prompt contains four key components: (1) user reference images (both preferred and non-preferred), (2) two candidate images, (3) meta-information including the question prompt, visual preferences and non-preferences, and the correct answer, and (4) system prompt along with correct response examples. Based on this prompt, Claude 3.7 follows a \textit{predict-then-assess} structure. It first predicts the user's visual preference profile from the reference images, which serves as a criterion for assessment. It then generates a CoT-style analysis that includes multi-dimensional, interpretable scoring and assessments for the candidate images, and finally outputs the final result. The response template is shown in Fig. \ref{personalized_cold_data}(d).
To ensure annotation quality, we further  filter out any illogical or incorrect CoT-style responses, such as those with inconsistent reasoning or a mismatch between the prediction and ground-truth. Finally, we curate PreferImg-CoT, a large-scale CoT-style dataset with 60,000 user samples based on the proposed PreferImg dataset, which serves as the foundation for our subsequent cold-start SFT.

\vspace{-2mm}
\subsection{Training Strategy}
\label{sec3.3}
We employ a two-stage training strategy to elicit and incentivize the model's structured reasoning capabilities. First, we conduct a cold-start supervised fine-tuning (SFT) phase to teach the model how to perform  structured reasoning. Then, we employ reinforcement learning (RL)-based post-training to encourage the model to explore more reasonable assessment paths and enhance its generalization.

\begin{figure}[t]
\vspace{-5mm}
  \setlength{\abovecaptionskip}{0pt}
  \setlength{\belowcaptionskip}{-15pt}
    \captionsetup{skip=2pt} % 仅对当前图片生效
  \centering
   \includegraphics[width=1\linewidth]{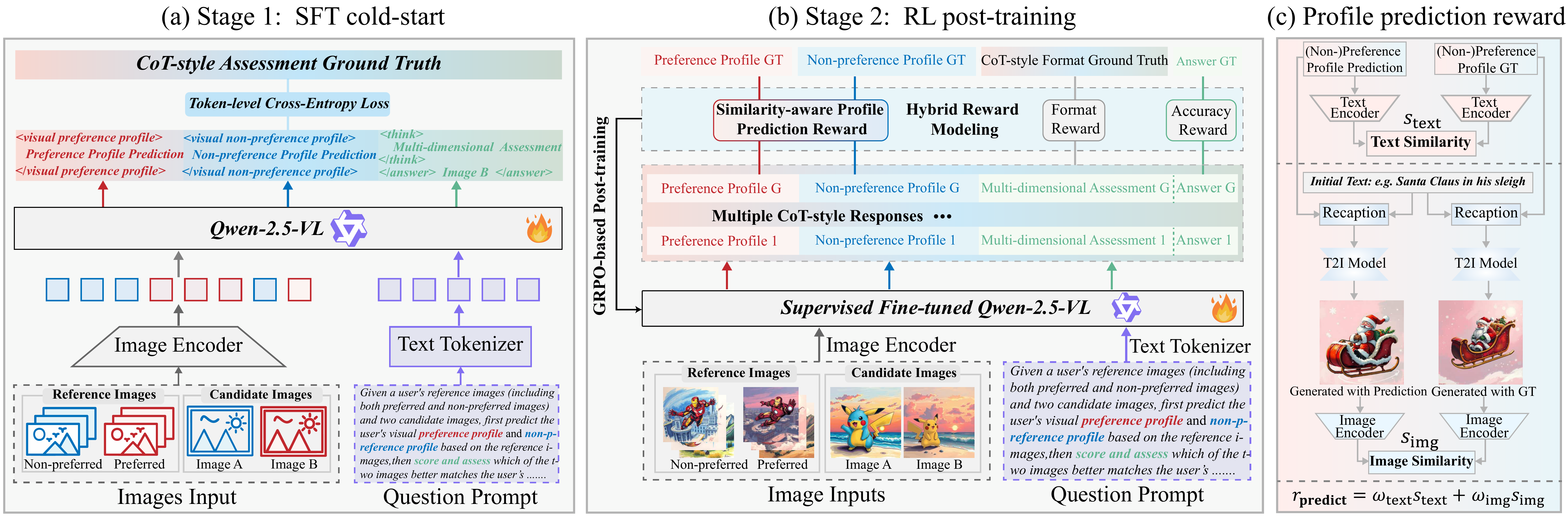}
   \caption{Illustration of training strategy and proposed prediction reward: (a) Cold-start SFT to teach  structured reasoning; (b) RL-based post-training to explore more reasonable assessments and enhance model generalization. (c) Similarity-aware prediction reward for better preference profile prediction.}
   \label{Train_stages}

\end{figure}

\textbf{Stage 1: Supervised Fine-tuning for Cold-start Initialization.} We utilize Qwen2.5-VL-7B \citep{bai2025qwen2} as the baseline model and conduct a SFT with the proposed CoT-style dataset PreferImg-CoT, as shown in Fig. \ref{Train_stages}(a). During the cold-start phase, the model learns to first predict the user's preference profiles based on reference images and then leverages the predicted profiles as criteria to  produce interpretable, multi-dimensional scores and assessments for candidate images.  We train the model using a standard autoregressive language modeling objective and apply a token-level cross-entropy loss to provide strong supervision throughout the generation process: 
\begin{equation}\mathcal{L}_{SFT}(\theta):=-\mathbb{E}_{(x,y)\thicksim\mathcal{D}_{\mathrm{CoT}}}\sum_{t=1}^T\log P\left(y_t\mid x,y_{<t};\theta\right),\end{equation}
where $\mathcal{D}_{\mathrm{CoT}}$ is our  PreferImg-CoT, and $(x,y)$ is the input query and CoT-style target response.

\textbf{Stage 2: GRPO-based Reinforcement  Learning for Post-training.}
Although cold-start endows the model with structured  reasoning capabilities, its supervised training paradigm limits the exploration of diverse reasoning paths and restricts  generalizability. To explore more reasonable assessments and enhance generalizability, we employ the Group Relative Policy Optimization (GRPO)-based reinforcement learning for exploration-driven post-training, as shown in Fig. \ref{Train_stages}(b). Unlike traditional policy optimization methods \citep{schulman2017proximal},  GRPO efficiently generates diverse reasoning responses by optimizing policy gradients on a sample group without requiring a critic model.

\textit{Group Relative Policy Optimization.} The MLLM generates a group of $G$ CoT-style outputs $\{o_1,o_2,\ldots,o_G\}$ for each input query $x$ from current policy $\pi_{\theta}$.  Each output contains preference profiles prediction, multi-dimensional assessment and a final answer. For each $o_i$, we compute a scalar reward $r_i$, and normalize these rewards to estimate its group-relative advantage $A_i$: 
\begin{equation}A_i=\frac{r_i-\max\left(\{r_1,r_2,\cdots,r_G\}\right)}{\operatorname{std}\left(\{r_1,r_2,\cdots,r_G\}\right)}
\end{equation}
Then, the GRPO training objective is as follows:
\vspace{-2mm}
\begin{align}
\mathcal{J}_{GRPO}(\theta) = & \mathbb{E}_{\substack{x \sim P(X), \{o_i\}_{i=1}^G \sim \pi_{\theta_{\text{old}}} (O \mid x)}} \biggl[ \frac{1}{G} \sum_{i=1}^G \min \biggl(\frac{\pi_\theta(o_i \mid x)}{\pi_{\theta_{\text{old}}}(o_i \mid x)} A_i,\\
& \operatorname{clip}\left(\frac{\pi_\theta(o_i \mid x)}{\pi_{\theta_{\text{old}}}(o_i \mid x)}, 1-\varepsilon, 1+\varepsilon\right) A_i\biggr) \notag  - \beta \mathbb{D}_{KL}(\pi_\theta \| \pi_{\mathrm{SFT}}) \biggr]
\end{align}

where $\varepsilon$ and $\beta$ are hyperparameters, and $\pi_{\mathrm{SFT}}$, $\pi_{\theta}$, and $\pi_{\theta_{\text{old}}}$ are the model after cold-start SFT, the optimized model and the old policy model. The KL divergence term $\mathbb{D}_{KL}(\pi_\theta \| \pi_{\mathrm{SFT}})$  promotes controlled policy updates, balancing exploration and stability throughout training.

\subsection{Similarity-aware Preference Profile Prediction Reward}
\label{sec3.4}

% \begin{figure}[t]
% \vspace{2mm}
%   \setlength{\abovecaptionskip}{0pt}
%   \setlength{\belowcaptionskip}{-15pt}
%   \centering
%    \includegraphics[width=0.99\linewidth]{Figures/Figure6_predictionreward.eps}
%    \caption{ }
%    \label{Train_stages}

% \end{figure}

% \begin{figure}[t]
% \vspace{2mm}
%   \setlength{\abovecaptionskip}{0pt}
%   \setlength{\belowcaptionskip}{-15pt}
%   \centering
%    \includegraphics[width=0.99\linewidth]{Figures/Figure6_predictionreward_1.eps}
%    \caption{ }
%    \label{Motivation}

% \end{figure}

%先讲偏好预测奖励函数的动机 配上图再讲怎么做

\begin{figure}[t]
  \setlength{\abovecaptionskip}{0pt}
  \setlength{\belowcaptionskip}{-15pt}
    \captionsetup{skip=2pt} % 仅对当前图片生效
  \centering
   \includegraphics[width=1\linewidth]{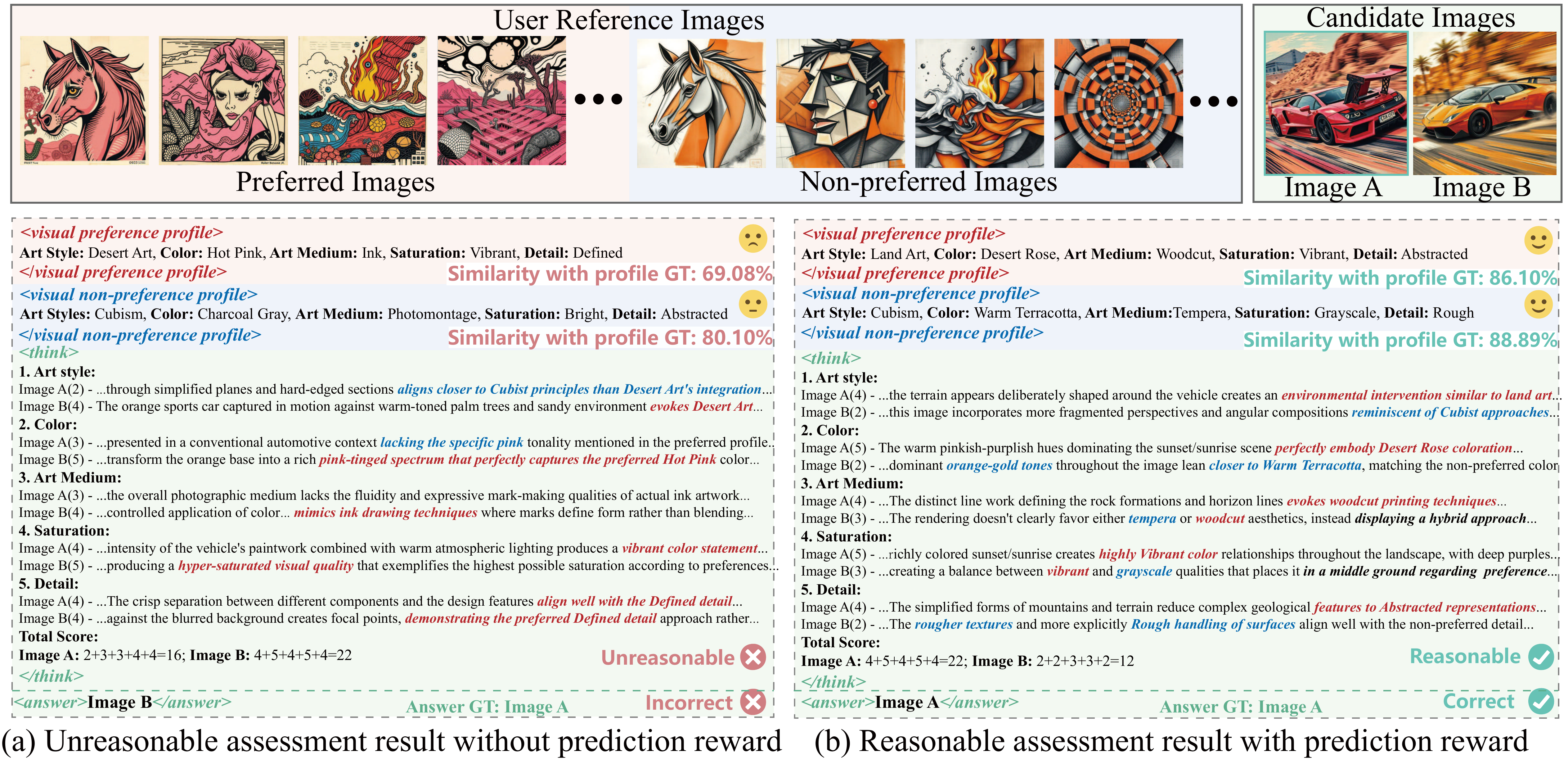}
   \caption{Effectiveness of the proposed prediction reward (PR). (a) The model without PR predicts profiles differed from GT, leading to unreasonable assessment and incorrect answer. (b) The model with PR predict the profiles closer to GT and provide more reasonable assessment with correct answer.}
   \label{Prediction Reward Effectiveness}

\end{figure}

We find that accurate profile prediction facilitates the exploration of reasonable assessments, motivating us to propose a prediction reward  to improve the accuracy of  profile predictions. Figure \ref{Prediction Reward Effectiveness}  illustrates a comparison of personalized assessment cases with and without the prediction reward . The results show that without the reward, the predicted profiles differ from the ground-truth (GT), leading to unreasonable assessments and an incorrect answer. Conversely, the model with the reward predicts the profiles closer to GT and provide more reasonable assessments with a correct answer.

% \begin{figure}[t]
% \vspace{-5.5mm}
%   \setlength{\abovecaptionskip}{0pt}
%   \setlength{\belowcaptionskip}{-15pt}
%     \captionsetup{skip=2pt} % 仅对当前图片生效
%   \centering
%    \includegraphics[width=1\linewidth]{Figures_pdf/Figure6_2.pdf}
%    \caption{Effectiveness of the proposed prediction reward (PR). (a) The model without PR predicts profiles differed from GT, leading to unreasonable assessment and incorrect answer. (b) The model with PR predict the profiles closer to GT and provide more reasonable assessment with correct answer.}
%    \label{Prediction Reward Effectiveness}

% \end{figure}

\textbf{Similarity-aware Prediction Reward Design.} To incentivize the model to  predict the user's preference profile more accurately, we propose a \textit{text-image} similarity-aware prediction reward in Fig. \ref{Train_stages}(c). Given a user's  personalized reference images, PreferThinker first predicts the user's visual preference profile $\hat{V}_{+}$ and non-preference profile $\hat{V}_{-}$.  First, we  measure the  prediction accuracy  via computing the \textit{text} semantic similarity $s_{\text{text}}$ between the predicted and ground-truth profiles:
\begin{equation}
  s_{\text{text}} = Sim_{\text{text}}(\hat{V}_{+}, V_{+}) + Sim_{\text{text}}(\hat{V}_{-}, V_{-})
  ,\end{equation}
where $Sim_{\text{text}}$ measures the similarity between two texts, $V_{+}$ and $V_{-}$  are ground-truth preference and non-preference profiles. We use SBERT  \citep{reimers2019sentence} to compute the text similarity.

Second, considering that image and text are the two sides of the same coin and should be tightly coupled. Therefore, we also measure the prediction accuracy via computing the \textit{image} similarity $s_{\text{img}}$ between the generated images based on the predicted and ground-truth visual preference profiles. Specifically, given an initial text $T_{\text{initial}}$,  we recap it by incorporating the predicted and ground-truth visual preference profiles, and then generate images using T2I model \citep{flux2024}:
\begin{equation}
\begin{aligned}
  \hat{I}_{+} &= {T2I}(R(T_{{\text{initial}}}, \hat{V}_{+})), & I_{+} &= {T2I}(R(T_{{\text{initial}}}, V_{+})), \\
  \hat{I}_{-} &= {T2I}(R(T_{{\text{initial}}}, \hat{V}_{-})), & I_{-} &= {T2I}(R(T_{{\text{initial}}}, V_{-})),
\end{aligned}
\end{equation}
where $T2I$ is the \text{text-to-image} model, $R$ denotes the recaption process. Then, we compute the image similarity between the images generated with predictions and ground-truth:
\begin{equation}
  s_{\text{img}} = Sim_{\text{img}}(\hat{I}_{+},{I}_{+} ) + Sim_{\text{img}}(\hat{I}_{-},{I}_{-} )
,\end{equation}
where $Sim_{\text{img}}$ measures the similarity between two images. We use DreamSim \citep{fu2023dreamsim} to compute the image similarity. Finally, the profile prediction reward function is a weighted combination of two similarity measures, designed to balance their contributions:
\begin{equation}
  r_{\mathbf{predict}} = w_{\text{img}}s_{\text{img}} + w_{\text{text}}s_{\text{text}}
,\end{equation}
where $w_{\text{img}}$ and $w_{\text{text}}$ are hyperparameters that balance the contribution of each similarity.

\textbf{Hybrid Reward Modeling.} Apart from considering the prediction reward ${r_{\mathbf{predict}}}$, we  utilize a format correct reward ${r_{\mathbf{format}}}$ and an assessment accuracy reward ${r_{\mathbf{accuracy}}}$ to encourage the model to produce the correct response format and assessment results:
\begin{itemize}[leftmargin=10pt]
    \item \textit{Format Correctness Reward} (${r_{\mathbf{format}}}$). The format reward ${r_{\mathbf{format}}}$ is awarded 1 when the response follows the required structure: 1) the predicted preference and non-preference profiles are enclosed within the \texttt{<visual preference profile>} and \texttt{<visual non-preference profile>} tags, respectively, 2) the multi-dimensional assessment is placed in the \texttt{<think>} tags, 3) the final result is contained in the  \texttt{<answer>} tags; otherwise, the ${r_{\mathbf{format}}}$  is set to 0.
    \setlength{\itemsep}{0pt}
    \item \textit{Assessment Accuracy Reward} (${r_{\mathbf{accuracy}}}$). Given the ground truth of Assessment result $E$ and the produced assessment result $\hat{E}$. The reward ${r_{\mathbf{accuracy}}}$ is set to 1 when  $\hat{E}= E$, and 0 otherwise.
\end{itemize}
Furthermore, during the RL-based post-training, we utilize a weighted combination to balance these reward functions. The total reward $r$ for a response is weighted sum of the three rewards:
\begin{equation}r= w_pr_{\mathbf{predict}} + w_fr_{\mathbf{format}} + w_ar_{\mathbf{accuracy}}
\end{equation}
where $w_p$, $w_f$ and $w_a$ are hyperparameters that balance the contribution of each reward.

\begin{wraptable}{r}{0.6\textwidth}
\vspace{-4pt}
\caption{\small Quantitative comparison of assessment accuracy with existing methods on the proposed  PreferImg and real user dataset PickaPic. SP denotes that the user's reference image reflects single preference, while MP is multiple preferences. $\boldsymbol{\dagger}$ denotes the dataset's  labels are for \textit{general preference assessment}. $*$ denotes the methods that support reference images as input for in-context learning. The \textbf{\textcolor{tablered}{best}} and \textbf{\textcolor{tableblue}{second-best}} performances are highlighted.}
\centering
\footnotesize
\setlength\tabcolsep{1.1pt}
\renewcommand\arraystretch{0.98}
\vspace{-8pt}
\resizebox{0.6\textwidth}{!}{ % 将宽度设置为与wraptable一致
\begin{tabular}{>{\raggedright\arraybackslash}p{4cm}|c|c c |c c c|c}
% \thickhline
\specialrule{1pt}{0pt}{0pt} % 设置第一条水平线的粗细
% \rowcolor[HTML]{f8f9fa}
% \multicolumn{1}{c||}{Method} & \#Param & \makecell{PreferImg \\ (Seen)}& \makecell{PreferImg \\ (Unseen)} & \makecell{PickaPic \\ (Unseen)} & \makecell{PickaPic \\ (Unseen)} & \makecell{PickaPic \\ (Unseen)}\\
\multicolumn{1}{c|}{\multirow{3}{*}{Method}} & \multicolumn{1}{c|}{\multirow{3}{*}{\#Param}} & \multicolumn{2}{c|}{\textbf{Seen}} &  \multicolumn{3}{c|}{\textbf{Unseen}} & {\multirow{3}{*}{\textbf{Avg}}} \\
\cline{3-7}
& & \makecell{PreferImg\\(SP)}&  \makecell{PreferImg\\(MP)}&  \makecell{PreferImg\\(SP)}&  \makecell{PreferImg\\(MP)} & PickaPic$\boldsymbol{^\dagger}$ &\\
\specialrule{1pt}{0pt}{0pt} % 设置第一条水平线的粗细
% \hline
\rowcolor[HTML]{EBF6F5}
\multicolumn{8}{c}{\textit{CLIP-based Models}} \\
PickScore~\pub{NIPS2023} &986M &49.6 & 48.4 &51.2&56.4 & \textbf{\textcolor{tablered}{67.9}} & 54.7\\
ImageReward~\pub{NIPS2023} &478M & 52.2  &51.2 &49.0&57.2 &60.7 &54.1\\
HPSv2~\pub{ICCV2023} &2B &51.4 &50.0 &50.8 &57.6 &63.3 &54.6\\
CLIPScore~\pub{ICML2023} &428M &50.2 &54.0 &52.6 &52.0  &58.0 &53.4 \\
Aesthetics~\pub{arXiv2021} &304M &50.6 &48.0 &48.6&50.8  &48.6 &49.3 \\
CycleReward~\pub{ICCV2025} &477M &48.0 &50.8 &50.0 &56.0 &62.6 &53.5\\
% \hline
% \hline
\rowcolor[HTML]{FEEBED}
\multicolumn{8}{c}{\textit{MLLM-based Models}} \\
UnifiedReward~\pub{arXiv2025} &7B &49.0 &46.0   &50.6 &57.2  &61.7 &52.9\\
UnifiedReward-Think~\pub{NIPS2025} &7B&48.6 &47.2 &48.0&52.8 &60.5 &51.4\\
LLaVA-Reward~\pub{ICCV2025} &8.2B &53.2 & 51.6&52.4&58.4 &62.2 &55.6\\
ViPer$^*$~\pub{ECCV2024} &8B &92.4 &78.0 & \textbf{\textcolor{tableblue}{93.4}} &  80.0 &62.2 &81.2\\
% \hline
% \hline
\rowcolor[HTML]{E3F4FB}
\multicolumn{8}{c}{\textit{Open-Source MLLMs}} \\
Qwen2.5-VL-7B ({\textit{Base}})$^*$~ &7B &75.4 &62.0 &72.0 &64.8 &58.0 &66.4  \\
InternVL-3.5-8B$^*$~&8B &80.0 &64.0 &80.6 &65.6 & 56.0& 69.2 \\
GLM-4.1V-9B-Thinking$^*$~&9B &86.8 &73.2 &86.8  &72.0 &62.8 & 76.3\\
\rowcolor[HTML]{FEF9E7}
\multicolumn{8}{c}{\textit{Closed-Source MLLMs}} \\
Claude 3.7$^*$ &\textcolor{mygray}{-} &93.8 &\textbf{\textcolor{tableblue}{83.2}} &90.2 &\textbf{\textcolor{tableblue}{86.0}} &  64.9& \textbf{\textcolor{tableblue}{83.6}}\\
OpenAI-GPT-4o$^*$&\textcolor{mygray}{-} &\textbf{\textcolor{tableblue}{94.2}} &80.4 &92.2 & 85.2& \textbf{\textcolor{tableblue}{65.7}} & 83.5  \\
Doubao-1.5-vision-pro-32k$^*$ &\textcolor{mygray}{-}&92.2 &78.4 &90.0 &76.4 &63.8 & 80.2 \\
Gemini-2.5-flash$^*$ &\textcolor{mygray}{-}&85.2 &61.6 &83.6 &67.2 &61.1 &71.7  \\
\rowcolor[HTML]{D5F5E3}
\multicolumn{8}{c}{\textbf{\textit{Ours}}} \\
\textbf{PreferThinker}$^*$~\pub{This paper}& 7B &\textbf{\textcolor{tablered}{96.6}} &\textbf{\textcolor{tablered}{92.0}} &\textbf{\textcolor{tablered}{96.4}} & \textbf{\textcolor{tablered}{92.8}} 
& \textbf{\textcolor{tableblue}{65.7}}& \textbf{\textcolor{tablered}{88.7}}   \\
 \textit{vs. prev. SoTA}&\textcolor{mygray}{-}
 & \textbf{\textcolor{tablegreen}{+2.4}} & \textbf{\textcolor{tablegreen}{+8.8}}& \textbf{\textcolor{tablegreen}{+3.0}} & \textbf{\textcolor{tablegreen}{+6.8}} & \textbf{\textcolor{tablegreen}{-2.2}} & \textbf{\textcolor{tablegreen}{+5.1}}   \\ 
 \textit{vs. Base Model}&\textcolor{mygray}{-}
 & \textbf{\textcolor{tablegreen}{+21.2}} & \textbf{\textcolor{tablegreen}{+30.0}}& \textbf{\textcolor{tablegreen}{+24.4}} & \textbf{\textcolor{tablegreen}{+28.0}} & \textbf{\textcolor{tablegreen}{+7.7}} & \textbf{\textcolor{tablegreen}{+22.3}}   \\ 
\specialrule{1pt}{0pt}{0pt} % 设置第一条水平线的粗细
\end{tabular}
 }
\label{table_comparison}
\vspace{-13pt}
\end{wraptable}
\vspace{-3pt}

\begin{figure}[t]
\vspace{-7.5mm}
  \setlength{\abovecaptionskip}{0pt}
  \setlength{\belowcaptionskip}{-15pt}
    \captionsetup{skip=2pt} % 仅对当前图片生效
  \centering
   \includegraphics[width=0.98\linewidth]{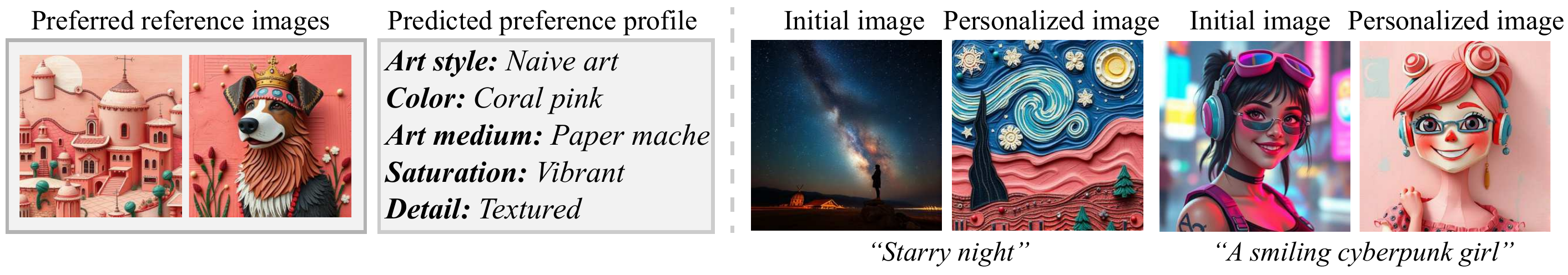}
   \vspace{-1mm}
   \caption{Personalized image generation with the predicated preference profile.}
   \label{Generation with predicted profile}

\end{figure}

% \begin{figure}[t]
%     \vspace{-5mm}
%   \centering
%   \begin{adjustbox}{width=0.95\columnwidth}
% 	\begin{minipage}[t]{0.47\linewidth}
%     \captionsetup{skip=1.1mm} % 设置标签与图之间的距离为10pt
% 		\includegraphics[width=1\linewidth]{Figures_pdf/Figure7_1.pdf}
% 		\caption{}
% 		\label{Generation with predicted profile}%文中引用该图片代号
% 	\end{minipage}
%   \hfill
% 	\begin{minipage}[t]{0.47\linewidth}
%     \captionsetup{skip=1.1mm} % 设置标签与图之间的距离为10pt
% 		\includegraphics[width=1\linewidth]{Figures_pdf/Figure7_1.pdf}
% 		\caption{Personalized image generation with the predicated preference profile.}
% 		\label{EFF_SLRTR}%文中引用该图片代号
% 	\end{minipage}
  
% \end{adjustbox}
%   \vspace{-8mm}
% \end{figure}

%% file: Secs/experiments.tex
\section{Experimental Results}
\subsection{Experimental Settings}
\label{sec4.1}

\vspace{-3pt}
\textbf{Datasets.} We conduct experiments on the proposed  PreferImg and real user dataset PickaPic \citep{kirstain2023pick}. Based on the PreferImg, we construct a benchmark of 1,500 users, categorizing them into single-preference (SP) and multi-preference (MP) groups based on whether reference images reflect singular or multiple preferences. To evaluate generalization, we further divide the benchmark into seen and unseen subsets based on whether users' profiles are included in the training data. 
Since no real-user dataset exists for personalized preference assessment, we process PickaPic, originally collected from real user interactions for \textit{general preference assessment}, by grouping samples by user IDs, resulting in a user-specific benchmark with 894 user samples. \textit{Note that although the samples are grouped by users, \textbf{the assessment labels still reflect general preferences}, unlike PreferImg, which targets personalized preferences}.

\textbf{Implementation Details.}  PreferThinker is initialized from Qwen2.5-VL-7B and  trained on a cluster of 8 NVIDIA A100 GPUs. We utilize the LLaMA-Factory \citep{zheng2024llamafactory} for cold-start SFT, and train for only one epoch to prevent overfitting. After the SFT, we utilize the VLM-R1 framework \citep{shen2025vlm} for RL-based post-training. We set a learning rate of 1e-5 with 6 rollout samples per input. The reward weights  $w_p$, $w_f$ and $w_a$ are set to 0.7, 0.3 and 1.0. See Appendix \ref{sec:supp_training_details} for details.

\textbf{Baselines.} We compare PreferThinker with (1) CLIP-based  models: PickScore \citep{kirstain2023pick}, ImageReward \citep{xu2023imagereward}, HPSv2 \citep{wu2023human2}, CLIPScore \citep{radford2021learning}, Aesthetics \citep{schuhmann2021laion}, CycleReward \citep{bahng2025cycle}; (2) MLLM-based methods:, UnifiedReward \citep{wang2025unified} UnifiedReward-Think \citep{wang2025unifiedthink}, LLaVA-Reward \citep{zhou2025multimodal}, and ViPer \citep{salehi2024viper}; (3) Open-source MLLMs: Qwen2.5-VL-7B \citep{bai2025qwen2}, InternVL-3.5 \citep{wang2025internvl3}, GLM-4.1V-9B \citep{hong2025glm}; (4) Closed-source MLLMs: Claude 3.7, GPT-4o, Doubao-1.5-vision, and Gemini-2.5. For a fair comparison, we  categorize the methods based on whether support reference images as input for in-context learning.

\vspace{-8pt}
\subsection{Comparison results}
\label{sec4.2}

\textbf{Comparison on Seen User Data.}  
%我们将preferthinker与现有的方法在域内单偏好用户和多偏好用户数据上进行对比in table。可以观察到基于clip的方法和大部分的基于MLLM的偏好评估方法性能较差，这是因为他们专门针对通用偏好评估，缺乏对个性化偏好的建模。其中ViPer取得较为良好的效果，但是仍落后于我们的方法，这是因为ViPer未充分利用参考图像的先验信息。进一步，我们也与开源和闭源的MLLMS进行对比，可发现直接使用开源MLLMS难以处理个性化偏好评估任务，闭源的MLLMS表现较为良好，但是仍落后于我们的方法。相比之下，所提方法取得了最优评估精度，
%给个图 与qwen进行对比
In Table \ref{table_comparison},  we compare PreferThinker with existing methods on the seen user data of PreferImg. It is observed that CLIP-based and most MLLM-based methods perform poorly, since they lack modeling for personalized preferences. Though ViPer achieves relatively good results on the SP data, its accuracy decreases when handling the MP data,  and it lacks interpretability as it only outputs a numerical score.  Further comparisons with the open-MLLMs reveal that directly utilizing  middle-sized models  struggle with personalized preference assessment. In contrast, closed-source models achieve comparable performance, thanks to their powerful prior knowledge.   Overall, PreferThinker achieves the highest accuracy and  provides interpretable assessments and   scores. 

\textbf{Comparison on Unseen User Data.}
%我们进一步在preferImg没有见过的用户数据上与现有的方法进行对比in Table \ref{table_comparison}. 可以观察到我们的方法在没有见过的用户数据集上仍然取得了较好的评估精度。同时了，为了验证方法在真实用户数据上的泛化性，我们也在真实数据PickaPic\_v2上进行了对比。可以发现真实用户个性化偏好数据更加复杂且存在低质量数据，所以整体精度较低，但我们的方法仍然取得了最优的评估精度，验证了所提方法对没有见过的数据的泛化性。
To validate generalization, we  compare performance on unseen users in Table \ref{table_comparison}.  PreferThinker  maintains the highest accuracy on  PreferImg,  revealing strong generalization. However, nearly all methods perform poorly on the real user benchmark, PickaPic, due to its complexity. Though we group  PickaPic by user ID and assign user-specific reference images, \textit{the labels still reflect general preferences}. Thus, methods designed for general preferences perform better on PickaPic. PickScore achieves the best performance since PickaPic serves as a seen dataset for it, while our method ranks second on unseen PickaPic data, further confirming its generalizability. \textit{Please refer to Appendix \ref{sec:supp_visualization_real_user} for CoT-style assessment visualization of real users.}

\vspace{-8pt}
\subsection{Ablation Study and Discussion}
\label{sec4.3}

\textbf{Effectiveness of Cold-start SFT and RL Post-training.} Table \ref{Ablation} demonstrates the critical roles of both cold SFT and RL training phase. Removing either phase leads to a notable degradation in model performance, confirming their importance for accurate preference assessment. Moreover, the use of RL after SFT significantly improves the model's performance on unseen user data, particularly for multi-preference (MP) users, revealing its effectiveness in enhancing model generalization.

\begin{table}[t]
  \vspace{-20pt}
    \centering
      % \begin{adjustbox}{width=0.98\columnwidth}
    \setlength{\tabcolsep}{1.2pt} % Reduce column separation
    
    % First table in a minipage
    \begin{minipage}[t]{0.59\textwidth}\scriptsize
        \centering
        \caption{Ablation study of PreferThinker. \textit{Ass.}: Assessment accuracy. \textit{Pred.}: Profile prediction accuracy.}
        \vspace{-8pt}
          \renewcommand\arraystretch{1.1}
        \resizebox{\linewidth}{!}{
            \begin{tabular}{>{\raggedright\arraybackslash}c c c c|c c|c c|c c|c c}
\specialrule{1pt}{0pt}{0pt} % 设置第一条水平线的粗细
            \multicolumn{1}{c}{\multirow{2}{*}{Base}} & \multicolumn{1}{c}{\multirow{2}{*}{SFT}} & \multicolumn{1}{c}{\multirow{2}{*}{RL}} & \multicolumn{1}{c|}{\multirow{2}{*}{PR}} & \multicolumn{2}{c|}{Seen-SP} & \multicolumn{2}{c|}{Unseen-SP} & \multicolumn{2}{c|}{Seen-MP} & \multicolumn{2}{c}{Unseen-MP} \\
            \cline{5-12}
            &&&& \textit{Ass.} &\textit{Pred.}&  \textit{Ass.} &\textit{Pred.}&\textit{Ass.}&\textit{Pred.}&  \textit{Ass.}&\textit{Pred.}\\
\specialrule{1pt}{0pt}{0pt} % 设置第一条水平线的粗细
            \ding{51} & & & &75.4 & 70.4 &72.0 &70.88 & 62.0&71.1  &64.8 &71.1\\
            \ding{51}& \ding{51} & & & 92.0&84.2 &91.8 &85.3 &81.2 & 73.9 &81.6 &74.2\\
            \ding{51}& & \ding{51} & & 89.6&79.6 &91.0 &80.3 &78.4 &76.9 &77.2 &73.5 \\
            \ding{51}& \ding{51}& \ding{51}& &94.9 &84.1 &95.2 &85.2 & 90.4& 72.5  &89.2 &74.1 \\
            \rowcolor[HTML]{F2F2F2}
            \specialrule{1pt}{0pt}{0pt} % 设置第一条水平线的粗细
            \ding{51}& \ding{51}  & \ding{51}&\ding{51} &\textbf{96.6} & \textbf{85.4}&\textbf{96.4} &\textbf{86.2} & \textbf{92.0} & \textbf{78.6} &\textbf{92.8} &  \textbf{80.7}\\
            \specialrule{1pt}{0pt}{0pt} % 设置第一条水平线的粗细
            \end{tabular}
          \label{Ablation}
        }
    \end{minipage}%
    \hfill
    % Second table in a minipage
    \begin{minipage}[t]{0.38\textwidth}\scriptsize
        \centering
        \caption{Performance and speed of PreferThinker without \textit{explicit} CoT.}
        \vspace{-8pt}
        \renewcommand\arraystretch{1.215}
        \setlength\tabcolsep{0.7pt}
        \large
        \resizebox{\linewidth}{!}{
            \begin{tabular}{c|cc|c}
            % \toprule
            % \toprule
            \specialrule{1.8pt}{0pt}{0pt} % 设置第一条水平线的粗细
        \multicolumn{1}{c|}{\multirow{2}{*}{Method}} & \multicolumn{2}{c|}{\textit{Accuracy}$\uparrow$ }  & \multicolumn{1}{c}{\multirow{2}{*}{\textit{Speed}$\downarrow$ }} \\
        \cline{2-3}
              & PreferImg & PickaPic$\boldsymbol{^\dagger}$\\
            \specialrule{1.8pt}{0pt}{0pt} % 设置第一条水平线的粗细
            ViPer & 86.0 &62.2 & \textbf{\textcolor{tableblue}{2.19 s}}  \\
            OpenAI-GPT-4o & 88.0 &\textbf{\textcolor{tablered}{65.7}} & 12.92 s \\
            Claude 3.7 & \textbf{\textcolor{tableblue}{88.3}}& 64.9& 18.22 s \\
            Doubao-1.5-vision  & 84.3& 63.8& 5.39 s \\
                      \rowcolor[HTML]{F2F2F2}
            \specialrule{1.8pt}{0pt}{0pt} % 设置第一条水平线的粗细
            \textbf{Ours (w/o CoT)} & \textbf{\textcolor{tablered}{93.2}}& \textbf{\textcolor{tableblue}{65.2}} & \textbf{\textcolor{tablered}{0.80 s}} \\
            \specialrule{1.8pt}{0pt}{0pt} % 设置第一条水平线的粗细
            \end{tabular}
            \label{performance and speed}
        }
    \end{minipage}
    % \end{adjustbox}
    \vspace{-3mm}
\end{table}

\begin{figure}[t]
% \vspace{-3mm}
  \setlength{\abovecaptionskip}{0pt}
  \setlength{\belowcaptionskip}{-15pt}
    \captionsetup{skip=2pt} % 仅对当前图片生效
  \centering
   \includegraphics[width=1\linewidth]{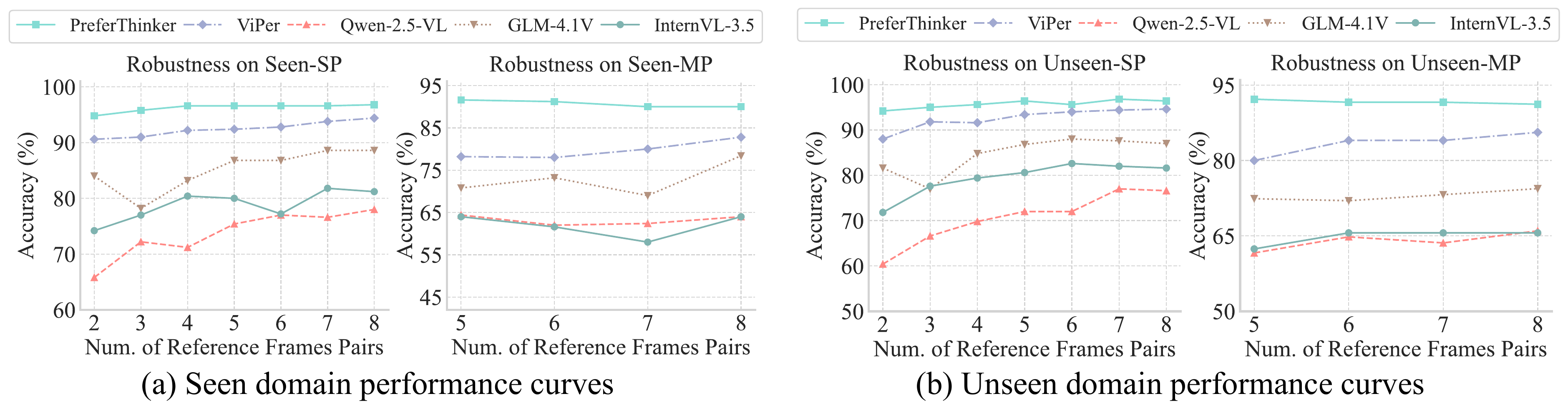}
   \caption{Robustness to number of personalized reference images on PreferImg.}
   \label{Robustness}
	\vspace{-7pt}
\end{figure}

%我们进一步探讨了SFT以及RL阶段对于偏好评估的重要性 in Table. 可以观察到去掉sft或者rl均会导致模型对于评估精度以及准则预测精度的下降，说明两个阶段均对于提升模型能力有重要作用。 另外，我们也观察到SFT阶段对于提升模型在域内数据上的表现更为重要，而RL阶段则对于提升模型在域外数据上的表现更为重要。

\textbf{How dose the Prediction Reward (PR) Facilitate Reasonable Assessment?} 
We further study the importance of the proposed prediction reward. Table \ref{Ablation} shows that adding the prediction reward effectively improves the preference profile prediction accuracy. Besides, Figure \ref{Prediction Reward Effectiveness} shows that incorporating the prediction reward enables accurate profile prediction, which  facilitates more reasonable assessment and correct answer, further demonstrating the effectiveness of the prediction reward.

\textbf{Personalized Generation with Predicted Preference Profile.} We further discuss how to utilize the predicated profile to generate personalized images in Fig. \ref{Generation with predicted profile}. We first predict a user's preference profile based on reference images. Then, we optimize the initial texts based on the predicated profile to generate  personalized images. The results show that the generated images align well with reference images, affirming the potential of the PreferThinker for facilitating personalized image generation. 

\textbf{Performance and  Speed of PreferThinker without Explicit CoT.} We study whether PreferThinker can implicitly use reasoning capabilities for direct preference assessment. Table \ref{performance and speed} shows that it still outperforms SOTA methods on PreferImg and achieves comparable accuracy on PickaPic even without explicit CoT, while delivering the fastest inference speed, further demonstrating its superiority.

\textbf{Robustness to Number of Reference Images.} To evaluate the robustness of PreferThinker, we use different number of reference images as user's prior information. Figure \ref{Robustness} shows that PreferThinker  outperforms other baselines across different numbers of reference images on both seen and unseen data, revealing its robustness in leveraging limited information for personalized preference assessment. 
%TODO 与 QWENVL intervl  Viper 以及 自己 进行曲线对比  做成曲线或者表格

%% file: Secs/appendix.tex
\section*{Appendix}
For a better understanding of the main paper, we provide additional details in this supplementary material, which is organized as follows:

\setcounter{figure}{0}
% 更改图表计数器格式为S1, S2, ...
\renewcommand{\thefigure}{S\arabic{figure}}
\renewcommand{\figurename}{Figure}

\setcounter{table}{0}
% 更改图表计数器格式为S1, S2, ...
\renewcommand{\thetable}{S\arabic{table}}
\renewcommand{\tablename}{Table}

%1.实验细节 sft的训练细节 以及 grpo的训练细节 参数
%2.冷启动数据集构建prompt 单偏好构建冷启动数据的prompt 多偏好构建冷启动数据的prompt
%3. 数据集细节
%3.可视化评估结果
%4.个性化生成结果
%6.工作缺陷以及未来方向

(\textbf{Section \ref{sec:supp_training_details}}) We provide more  details about the experimental settings.

\begin{itemize}
\item We provide more statistics details of the PreferImg test benchmark in Sec. \ref{sec:benchmark_details}.
\item We provide more training details for the SFT cold-start in Sec. \ref{sec:sft_details}.
\item We provide more training details for the  reinforcement learning post-training in Sec. \ref{sec:RL_details}.
\end{itemize}

(\textbf{Section \ref{sec:supp_dataset_details}}) We provide more details about the constructed dataset PreferImg.
\begin{itemize}
\item We provide more details of visual elements comprising the preference profile in Sec. \ref{sec:supp_elements}.
\item We show several examples of the proposed  PreferImg in Sec. \ref{sec:supp_examples}.
\end{itemize}

(\textbf{Section \ref{sec:supp_prompt_template}}) We provide additional details about our designed prompt templates.
\begin{itemize}
\item We provide prompt templates for generating CoT-style cold-start data with Claude 3.7 in Sec. \ref{sec:prompt_cold_start}.
\item We provide prompt templates using during SFT cold-start and RL post-training phase in Sec. \ref{sec:prompt_SFT_RL}.
\end{itemize}

 (\textbf{Section \ref{sec:supp_visualization}}) We present more visualization results of personalized image preference assessment on the proposed dataset PreferImg and real user dataset PickaPic.

\begin{itemize}
\item We show more visualization results on the PreferImg  where users's reference images reflect single preference in Sec. \ref{sec:supp_visualization_sp}.
\item We show more visualization results on the PreferImg dataset where users's reference images reflect multiple preferences in Sec. \ref{sec:supp_visualization_mp}.
\item We show more visualization results on  PickaPic to demonstrate our method's generalizability to real-world users in Sec. \ref{sec:supp_visualization_real_user}.
\end{itemize}

(\textbf{Section \ref{sec:supp_discussion}}) 
We conduct further discussions on the effectiveness and performance of proposed PreferThinker, along with limitation and future work.

\begin{itemize}
% \item We  analyze the performance and inference speed of PreferThinker with and without CoT-style reasoning  in Sec. \ref{sec:supp_performance_speed}.
\item We  investigate  the effectiveness of the cold-start SFT in Sec. \ref{sec:supp_cold_start_effectiveness}.
\item We  discuss the transferability of the PreferThinker towards assessment and scoring of single image in Sec. \ref{sec:supp_score_single_image}.
\item We study the effectiveness of PreferThinker to facilitate personalized image generalization in Sec. \ref{sec:supp_personalized_image_generation}.
\item We discuss the limitation of PreferThinker and the future work in Sec. \ref{sec:Limitation_and_Future_work}.
\end{itemize}

(\textbf{Section \ref{supp_use_llm}}) 
We  describe how we utilize LLMs to assist our work. (Use of LLMs)

\section{Experimental Settings}
\label{sec:supp_training_details}

\subsection{PreferImg Test Benchmark.}
\label{sec:benchmark_details}
To comprehensively evaluate the performance of personalized image preference assessment, we construct a test benchmark comprising 1500 user samples based on the proposed PreferImg dataset. We categorize the samples into single-preference (SP) and multi-preference (MP) groups based on whether reference images reflect singular or multiple preferences. To evaluate the generalization of PreferThinker, we further divide the benchmark into in-domain (ID)  and out-of-domain (OD) subsets based on whether users' profiles are included in the training data. The detailed statistics of the benchmark are provided in Table \ref{tab:benchmark_statistics}.

\subsection{Training Details of SFT Cold-Start Phase.}
\label{sec:sft_details}
We utilize Qwen2.5-VL-7B Instruct model as our base model and perform supervised fine-tuning using LLaMA-Factory \citep{zheng2024llamafactory}. We train the model on the proposed CoT-style cold-start dataset PreferImg-CoT for one epoch. The dataset includes 50,000 user samples with a single preference and 10,000 user samples with multiple preferences. we freeze the parameters of the vision tower and the multi-modal projector, and fine-tune the LLM. We employ DeepSpeed Zero-3 optimization strategy to handle the memory requirements of
large models. The training settings of SFT are detailed in the Table \ref{tab:training_settings}.

\begin{table}[t]
    \centering
    \footnotesize
        \caption{PreferImg test benchmark data categories and statistics}
    \resizebox{0.5\textwidth}{!}{ % 将宽度设置为与wraptable一致
    \begin{tabular}{c||c||c}

\hline
\rowcolor[HTML]{f8f9fa}
       \multicolumn{2}{c||}{Data Categories} &  Number \\
\hline
\hline
        {\multirow{2}{*}{In-Domain}} & Singe-preference & 500\\
         & Multi-preferences & 250  \\
         \hline
\hline
         {\multirow{2}{*}{Out-of-Domain}} & Singe-preference & 500 \\
         & Multi-preferences& 250  \\
\hline
\hline
       \multicolumn{2}{c||}{Total} &  1500 \\
\hline

    \end{tabular}
    }

    \label{tab:benchmark_statistics}
\end{table}

\begin{table}[t]
    \centering
        \caption{Training settings for cold-start SFT stage.}
    \resizebox{1\textwidth}{!}{ % 将宽度设置为与wraptable一致
    \begin{tabular}{cc||cc||cc}
\hline
\hline
        batch size & 1 & maximum gradient norm & 1 & precision & bf16 \\
        gradient accumulation & 2 & learning rate scheduler & cosine & epochs & 1 \\
        learning rate & 1e-6 & max length & 32768 & times & 21.1h \\
        optimizer & AdamW & deepspeed & zero2 & GPU & 8x A800 \\
        warm up ratio & 0.1 & weight decay & 0.0 & trainable module & LLM \\
\hline
\hline
    \end{tabular}
    }

    \label{tab:training_settings}
\end{table}

\subsection{Training Details of Reinforcement Learning Post-training Phase.}
\label{sec:RL_details}
 Following the cold-start SFT, we employ GRPO-based reinforcement learning for post-training. We utilize the VLM-R1 framework for this phase. For GRPO training, we use 10,000 user samples with single preference and another 10,000 user samples with multiple preferences. The number of reference frames pairs (preferred and non-preferred images) used in training is 5.  The training data of the phase does not include CoT-style reasoning, since we aim to incentivize the model to explore more reasonable assessment paths. The training settings of RL are detailed in the Table \ref{tab:grpo_hyperparameters}

\begin{table}[t]
    \centering
      \caption{Training settings for GRPO-based RL training stage.}
          \resizebox{1\textwidth}{!}{ % 将宽度设置为与wraptable一致
    \begin{tabular}{cc||cc||cc}
\hline
\hline
        batch size per device & 2 & num of rollout & 6 & precision & bf16 \\
        gradient accumulation & 2 & $\beta$ & 0.04 & epochs & 0.5 \\
        learning rate & 1e-6 & temperature & 0.9 & times & 73h \\
        optimizer & AdamW & deepspeed & zero3 & GPU & 6x A100 \\
        warm up ratio & 0.03 & weight decay & 0.01 & trainable module & LLM \\
\hline
\hline
    \end{tabular}
          }
    \label{tab:grpo_hyperparameters}
\end{table}

\section{Dataset Details}
\label{sec:supp_dataset_details}
%可视化一些数据集的例子
%把每个单词给可视化出来

\subsection{Visual Preference Elements of Preference Profile.}
\label{sec:supp_elements}
\begin{figure}[t]
%   \vspace{-6mm}
  \setlength{\abovecaptionskip}{0pt}
  \setlength{\belowcaptionskip}{-15pt}
  \captionsetup{skip=2pt} % 仅对当前图片生效
  \centering
   \includegraphics[width=0.97\linewidth]{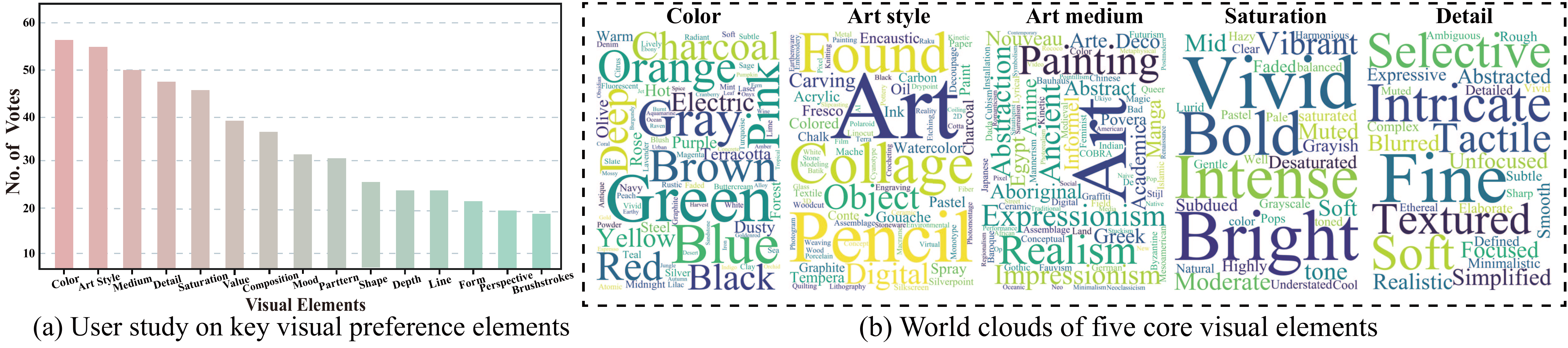}
   \caption{Key visual elements of preference profile. (a) User study result reveals that color, art style, art medium, saturation, and detail are voted five key elements representing the visual preference profile. (b)  World clouds show that each  element has a rich vocabulary associated with it.}
   \label{supp_criteria}
   %   \vspace{-1mm}
\end{figure}

To formalize a user's complex personalized visual preferences, we  identify 15 visual elements that most frequently appear in Lexica's \footnote{\href{https://lexica.art/}{https://lexica.art/}} text prompts and strongly influence user preference toward personalized image generation. We then conduct a user study with 100 participants, asking each to select the five most important visual elements as visual preference profile. The result in Fig \ref{supp_criteria}(a) reveals that \textit{art style}, \textit{color}, \textit{detail}, \textit{art medium} and \textit{saturation} are voted as the most representative visual elements for characterizing personalized preferences. To ensure preference profile diversity, we collect a rich vocabulary of related terms for each visual element, as shown in Fig. \ref{supp_criteria}(b), laying a solid foundation for constructing a large-scale and diverse dataset for personalized image preference assessment.

\begin{figure}[t]
  \setlength{\abovecaptionskip}{0pt}
  \setlength{\belowcaptionskip}{-15pt}
  \captionsetup{skip=2pt} % 仅对当前图片生效
  \centering
   \includegraphics[width=0.98\linewidth]{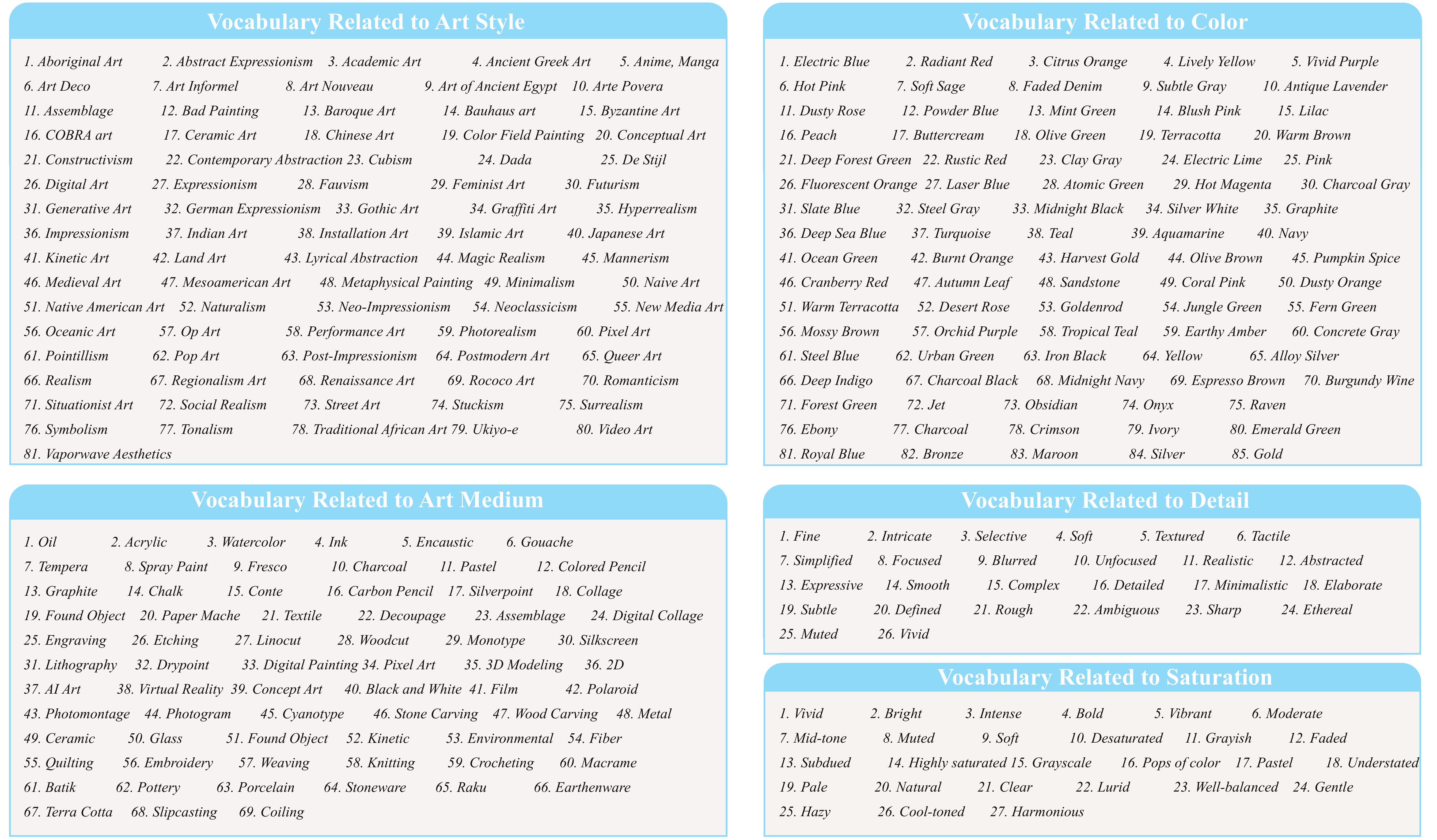}
   \caption{Vocabulary related to five core visual preference elements.}
   \label{supp_vocabulary}
\end{figure}

To ensure the diversity of personalized visual preference profiles, we collect an extensive set of vocabulary related to five key visual elements, as illustrated in Fig. \ref{supp_vocabulary}. Leveraging this lexicon, we construct a large-scale and rich set of visual preference profiles, laying a robust foundation for constructing the dataset PreferImg and CoT-style dataset PreferImg-CoT.

\subsection{Examples of the Proposed Dataset PreferImg.}
\label{sec:supp_examples}

\begin{figure}[t]
  \setlength{\abovecaptionskip}{0pt}
  \setlength{\belowcaptionskip}{-15pt}
  \captionsetup{skip=2pt} % 仅对当前图片生效
  \centering
   \includegraphics[width=0.98\linewidth]{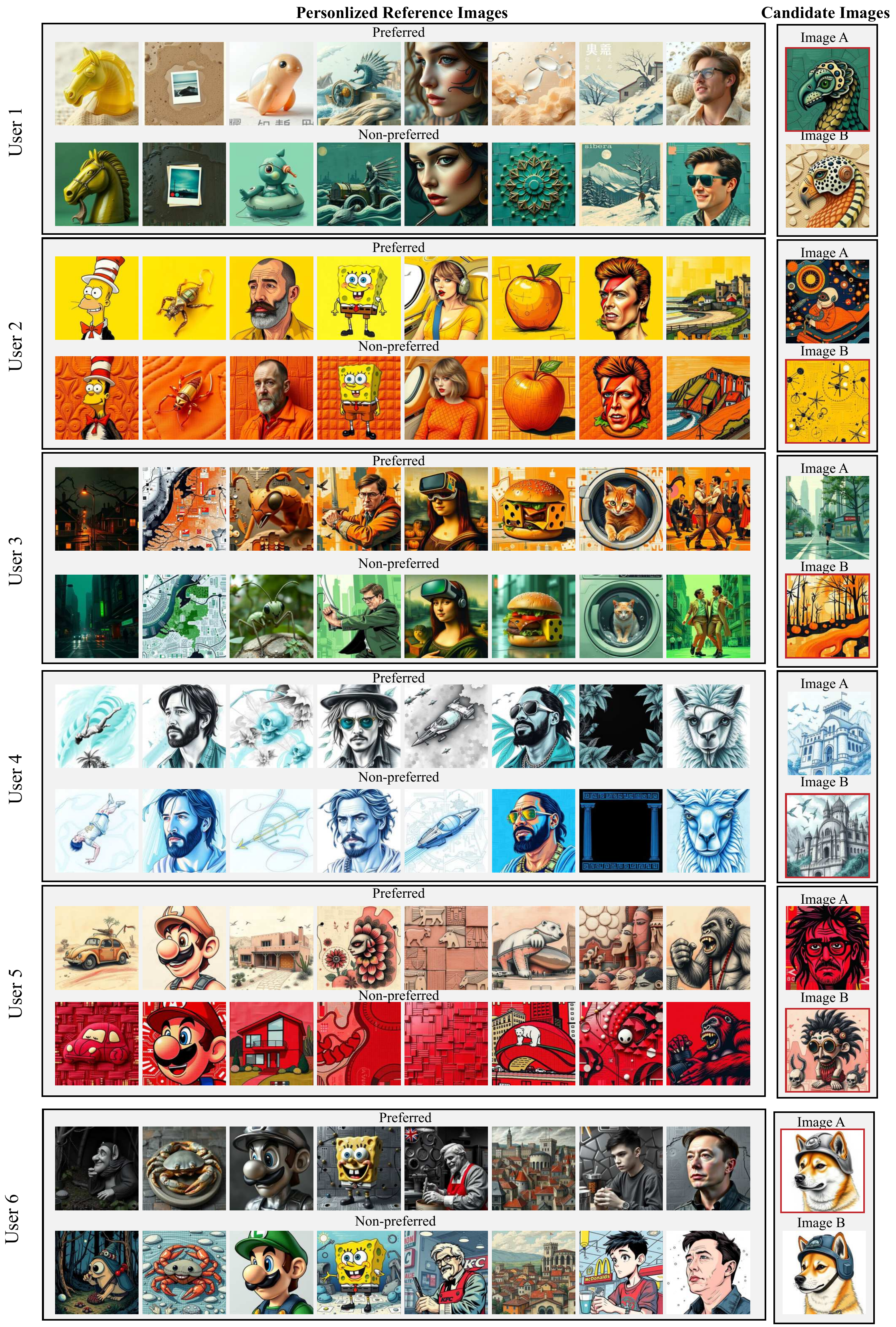}
   \caption{Examples of the proposed dataset PreferImg, where users' reference images reflect single preference. \textcolor{myred}{Red box} indicates preferred image in the candidate images.}
   \label{supp_sp_example}
\end{figure}

\begin{figure}[t]
  \setlength{\abovecaptionskip}{0pt}
  \setlength{\belowcaptionskip}{-15pt}
  \captionsetup{skip=2pt} % 仅对当前图片生效
  \centering
   \includegraphics[width=0.93\linewidth]{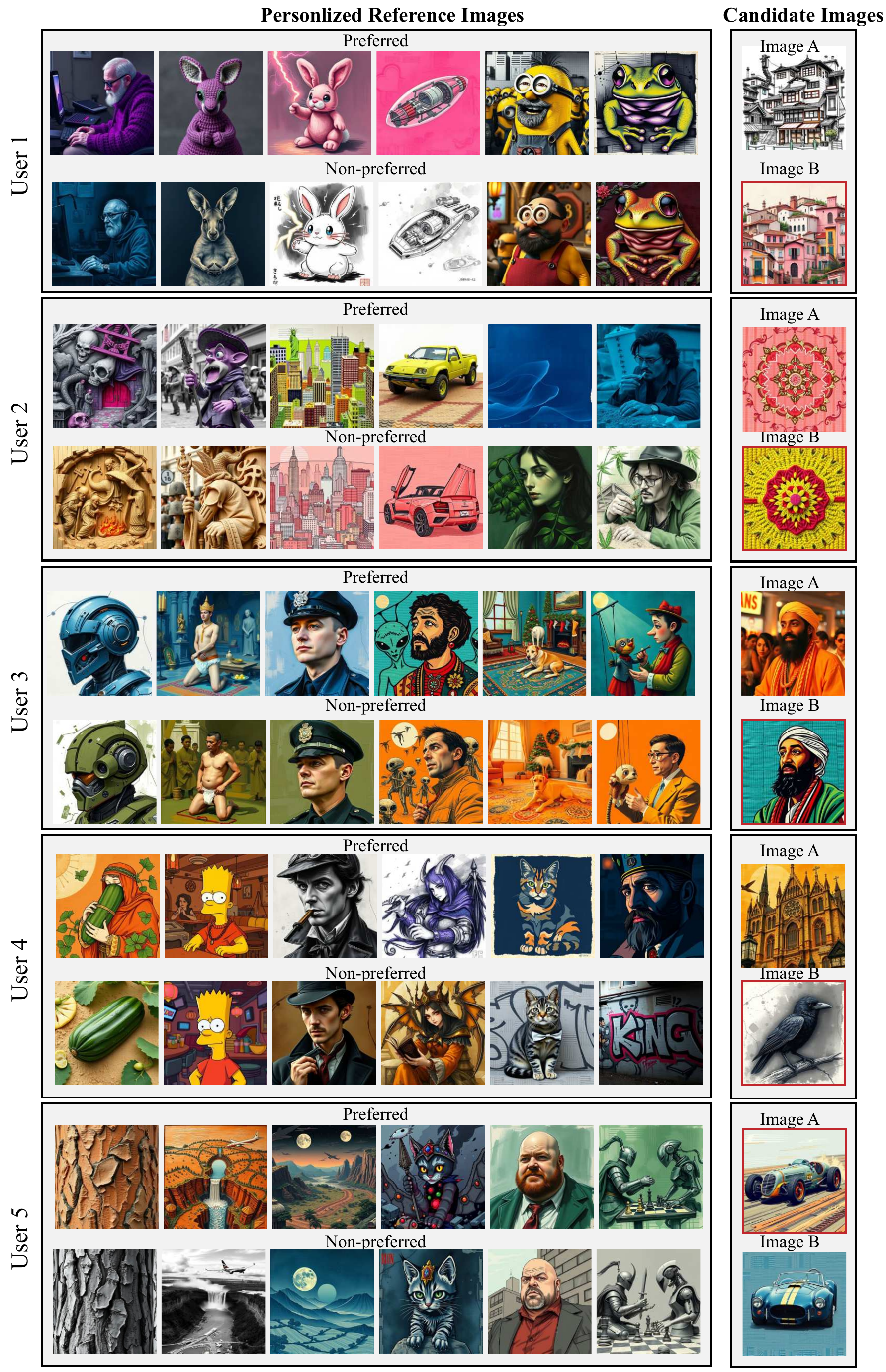}
   \caption{Examples of the proposed dataset PreferImg, where users' reference images reflect multiple preferences. \textcolor{myred}{Red box} indicates preferred image in the candidate images.}  
   \label{supp_mp_example}
\end{figure}

We visualize several user examples of the proposed dataset PreferImg in Fig. \ref{supp_sp_example} and Fig. \ref{supp_mp_example}. Figure. \ref{supp_sp_example} shows examples where users' reference images reflect single preference, while Fig. \ref{supp_mp_example} presents examples where users' reference images reflect multiple preferences. It is observed that our dataset encompasses a wide variety of image content and diverse visual preference profiles.

\section{Prompt Templates}
\label{sec:supp_prompt_template}

\subsection{Prompt  Templates for CoT-style Cold-start Data Generation.}
\label{sec:prompt_cold_start}

\begin{figure}[t]
  \setlength{\abovecaptionskip}{0pt}
  \setlength{\belowcaptionskip}{-15pt}
  \captionsetup{skip=2pt} % 仅对当前图片生效
  \centering
   \includegraphics[width=0.95\linewidth]{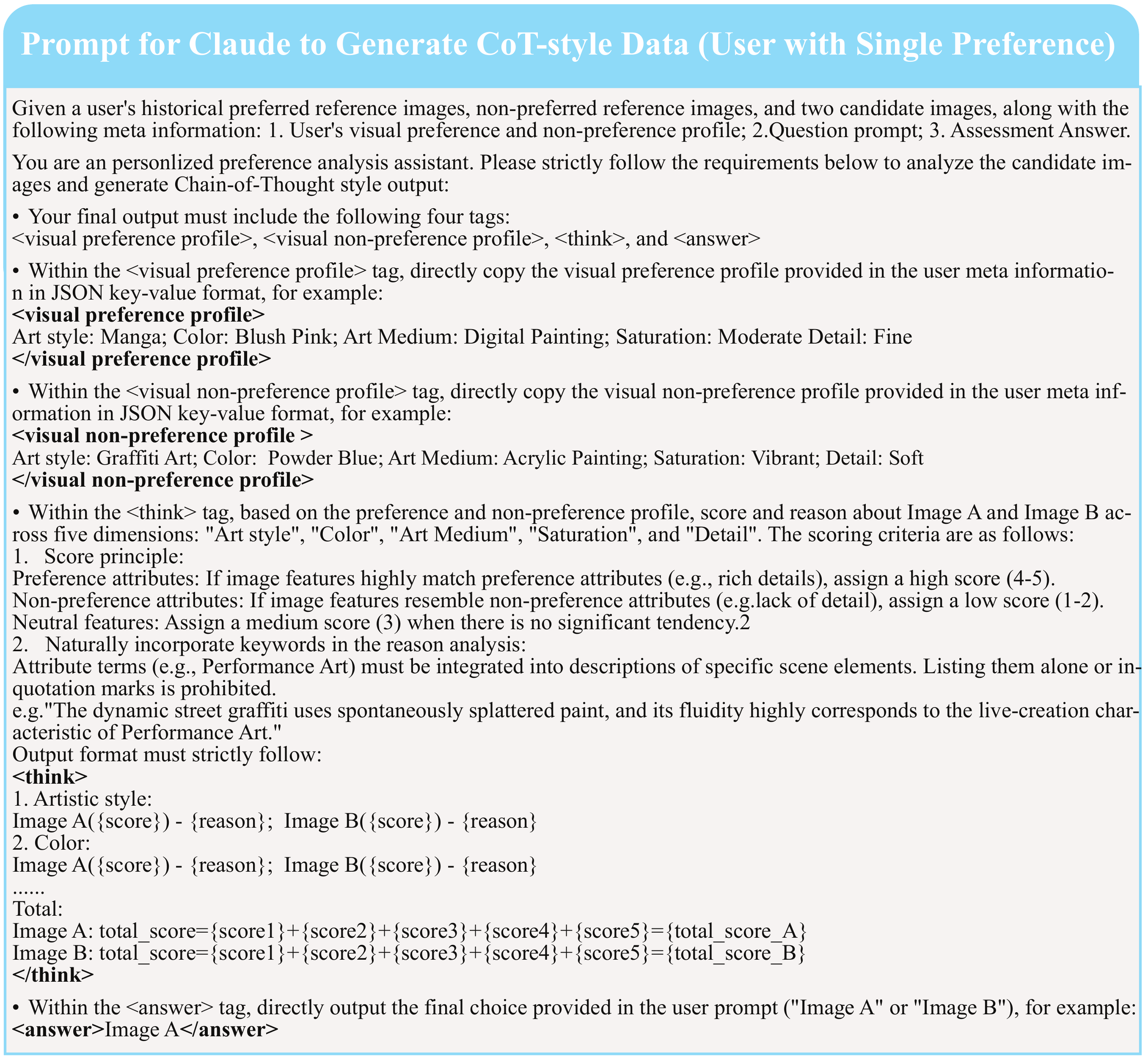}
   \caption{Prompt for Claude 3.7 to generate CoT-style assessment for users with single preference.}
   \label{supp_prompt_claude_sp}

\end{figure}

\begin{figure}[t]
  \setlength{\abovecaptionskip}{0pt}
  \setlength{\belowcaptionskip}{-15pt}
  \captionsetup{skip=2pt} % 仅对当前图片生效
  \centering
   \includegraphics[width=0.95\linewidth]{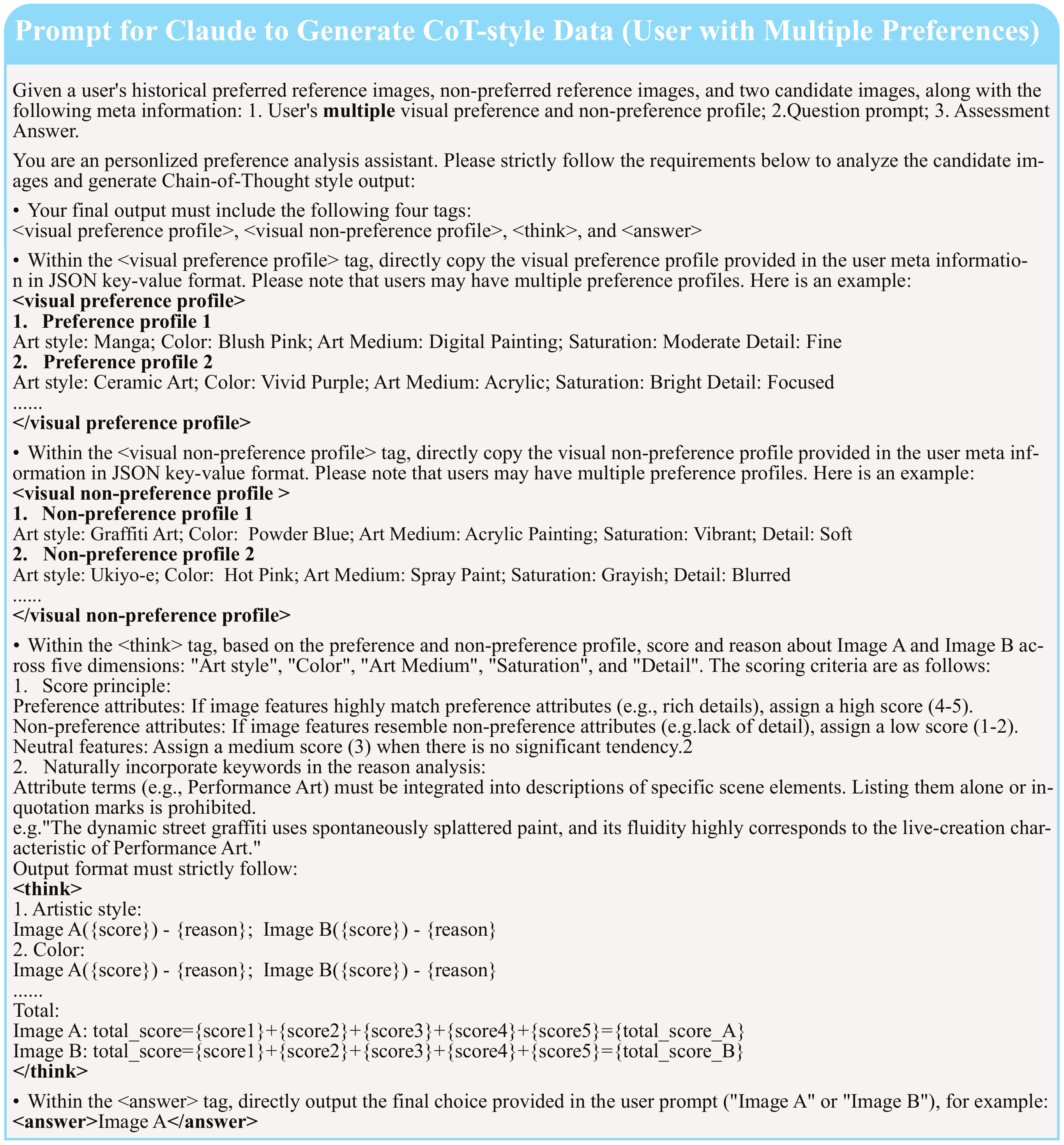}
   \caption{  Prompt for Claude 3.7 to generate CoT-style data for users with multiple preferences.}
   \label{supp_prompt_claude_mp}

\end{figure}

%为了使得Claude模型针对PreferImg数据集标注高质量的CoT风格的评估，我们设计详细的提示策略来引导claude模型结合用户的元信息对待选图像进行可解释性地打分和评估,其中元信息包括用户的视觉偏好和非偏好画像2.问题prompt3.评估答案。提示策略主要包含评估模板和准则。使评估输出遵循先预测后评估得框架：即先基于参考图像预测出用户的偏好画像，然后基于预测的偏好画像对待选图像进行多维度评估打分，最终根据总分结果选择用户偏好图像。用于生成CoT风格的评估答案的提示策略如图\ref{supp_prompt_claude_sp}和图\ref{supp_prompt_claude_mp}所示。图\ref{supp_prompt_claude_sp}展示了针对单偏好用户的提示策略，图\ref{supp_prompt_claude_mp}展示了针对多偏好用户的提示策略。我们使用Claude-2模型生成CoT风格的评估答案，并对生成结果进行人工筛选，最终得到5万单偏好用户样本和1万多偏好用户样本。
To enable the Claude 3.7 model to annotate high-quality, CoT-style assessments for the PreferImg dataset, we design a detailed prompting strategy that guides the Claude 3.7. This strategy directs Claude to use the meta information,including (1) user visual preference and non-preference profile, (2) question prompt, and (3) assessment answer to provide an explainable score and assessment for candidate images. The prompt mainly consists of an evaluation template and guidelines, follows a "predict-then-assess" structure. First, based on a reference image, the model predicts the user's preference profile. Then, using this predicted profile, it evaluates the candidate images across multiple dimensions and assigns interpretable scores. Finally, the model selects the user's preferred image based on the total scores. The prompts for generating the CoT-style assessment data are shown in Fig. \ref{supp_prompt_claude_sp} and Fig. \ref{supp_prompt_claude_mp}. Figure \ref{supp_prompt_claude_sp} shows the prompt for single-preference users, while Figure \ref{supp_prompt_claude_mp} shows the prompt for multi-preference users.

\subsection{Prompt Templates Using during SFT and RL Training.}
\label{sec:prompt_SFT_RL}
\begin{figure}[t]
  \setlength{\abovecaptionskip}{0pt}
  \setlength{\belowcaptionskip}{-15pt}
  \captionsetup{skip=2pt} % 仅对当前图片生效
  \centering
   \includegraphics[width=0.95\linewidth]{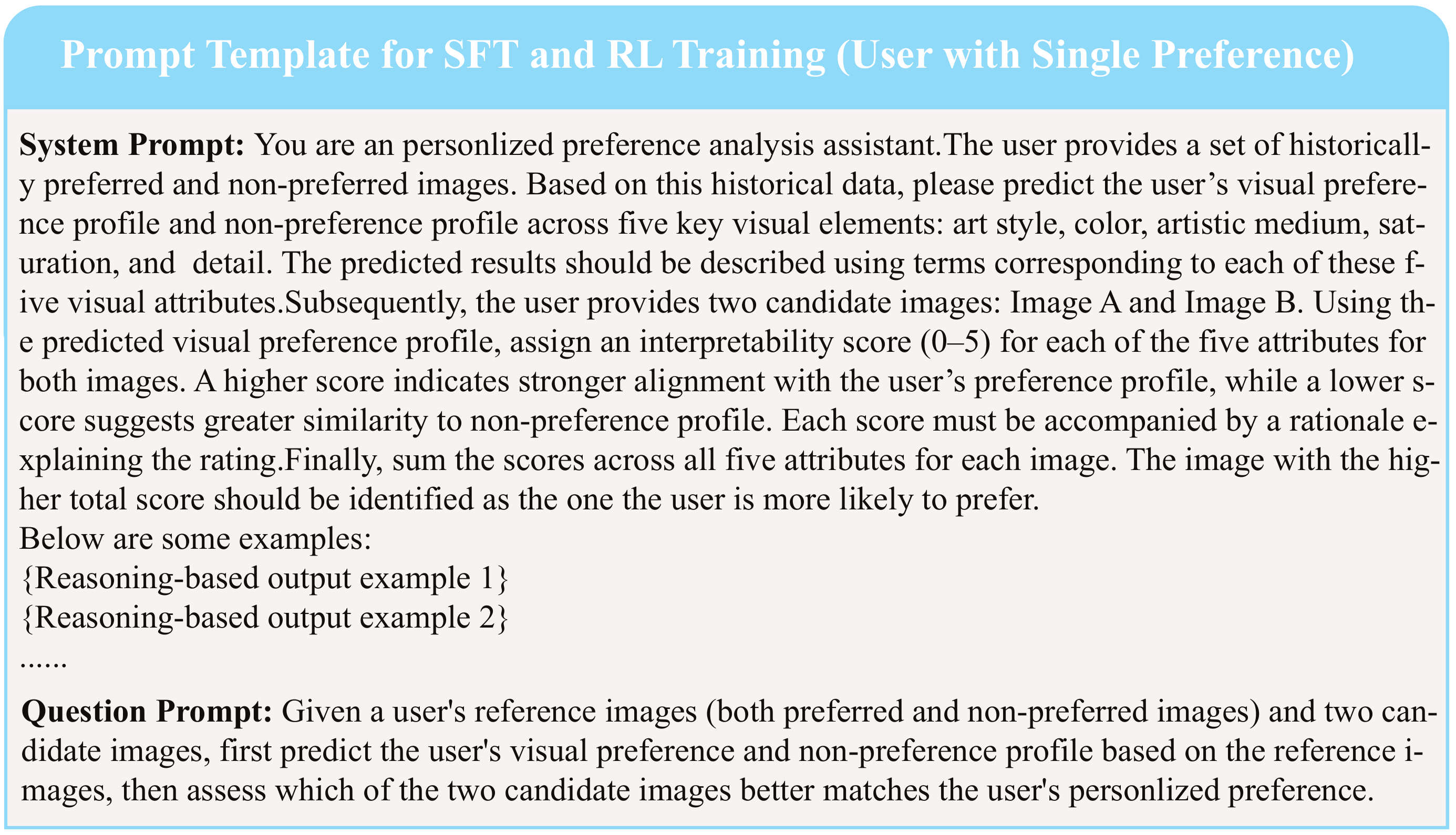}
   \caption{The SFT and RL prompt template for users with single preference}
   \label{supp_prompt_SFT_RL_sp}

\end{figure}

\begin{figure}[t]
  \setlength{\belowcaptionskip}{-15pt}
  \captionsetup{skip=2pt} % 仅对当前图片生效
  \centering
   \includegraphics[width=0.95\linewidth]{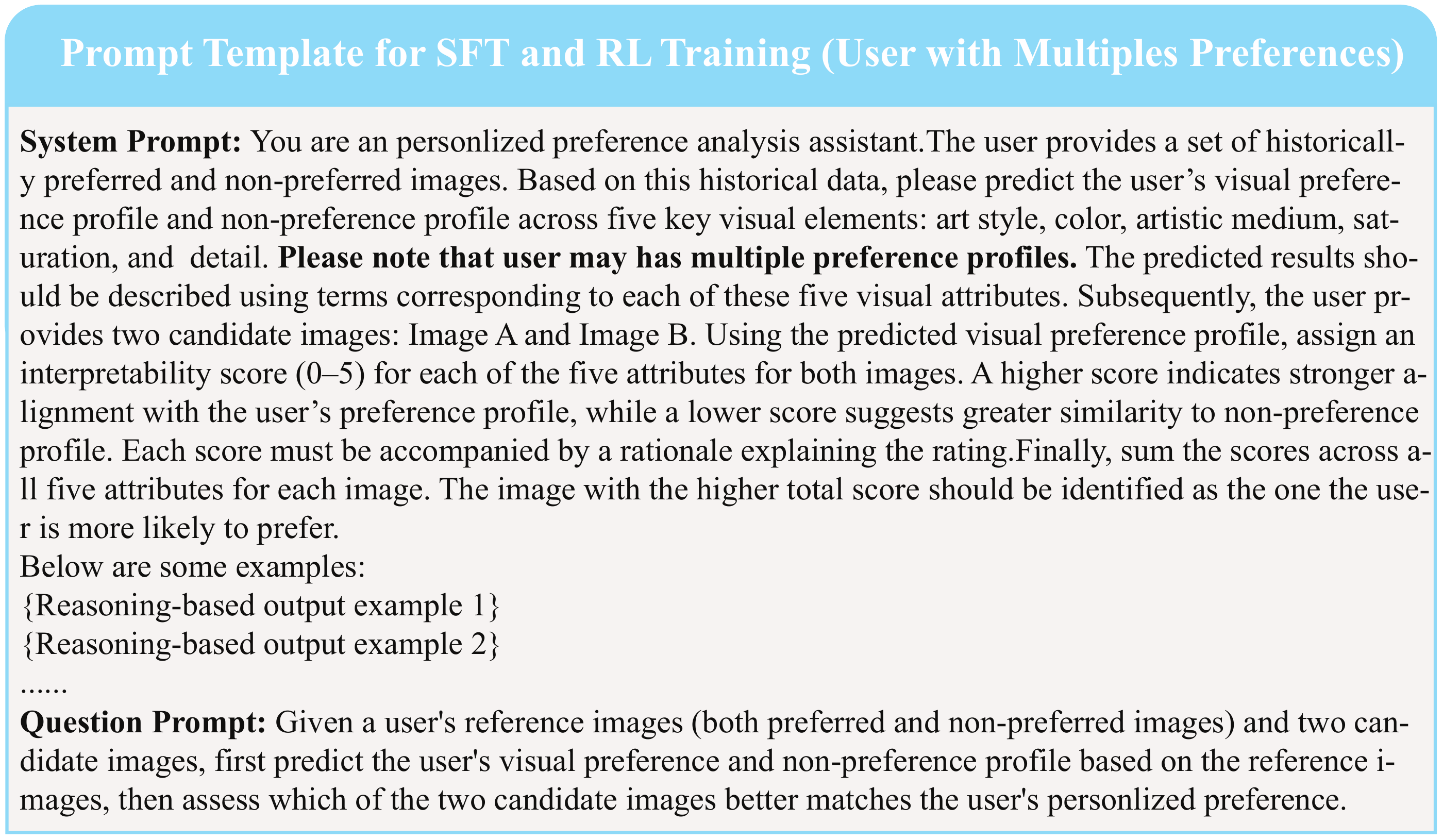}
   \caption{The SFT and RL prompt template for users with multiple preferences}
   \label{supp_prompt_SFT_RL_mp}

\end{figure}

% 我们提供了在冷启动sft和grpo rl训练时使用prompt模板，主要是包括系统提示和问题提示。这些提示模板主要描述了具体任务并给予正确的输出案列，引导模型能够按照正确的格式针对待选图像进行可解释性个性化偏好评估。 用于SFT和RL训练的提示策略如图\ref{supp_prompt_SFT_RL_sp}和图\ref{supp_prompt_SFT_RL_sp}所示。图\ref{supp_prompt_SFT_RL_sp}展示了针对单偏好用户的提示策略，图\ref{supp_prompt_SFT_RL_sp}展示了针对多偏好用户的提示策略。
We provide prompt templates for use during cold-start Supervised Fine-Tuning (SFT) and Group Policy Reinforcement Learning (GRPO RL) training. These templates, which include a system prompt and a question prompt, describe the specific task and provide correct output examples. This guides the model to perform explainable, personalized preference assessments for candidate images in the correct format. The prompt templates for SFT and RL training are shown in Fig. \ref{supp_prompt_SFT_RL_sp} and Fig. \ref{supp_prompt_SFT_RL_mp}. Figure \ref{supp_prompt_SFT_RL_sp} illustrates the prompt for single-preference users, while Figure \ref{supp_prompt_SFT_RL_mp} shows the prompt for multi-preference users.

\section{Visualization Results}
\label{sec:supp_visualization}

\subsection{Visualization Results on Users with Single Preference}
\label{sec:supp_visualization_sp}

\begin{figure}[b]
  \setlength{\belowcaptionskip}{-15pt}
  \captionsetup{skip=2pt} % 仅对当前图片生效
  \centering
   \includegraphics[width=0.99\linewidth]{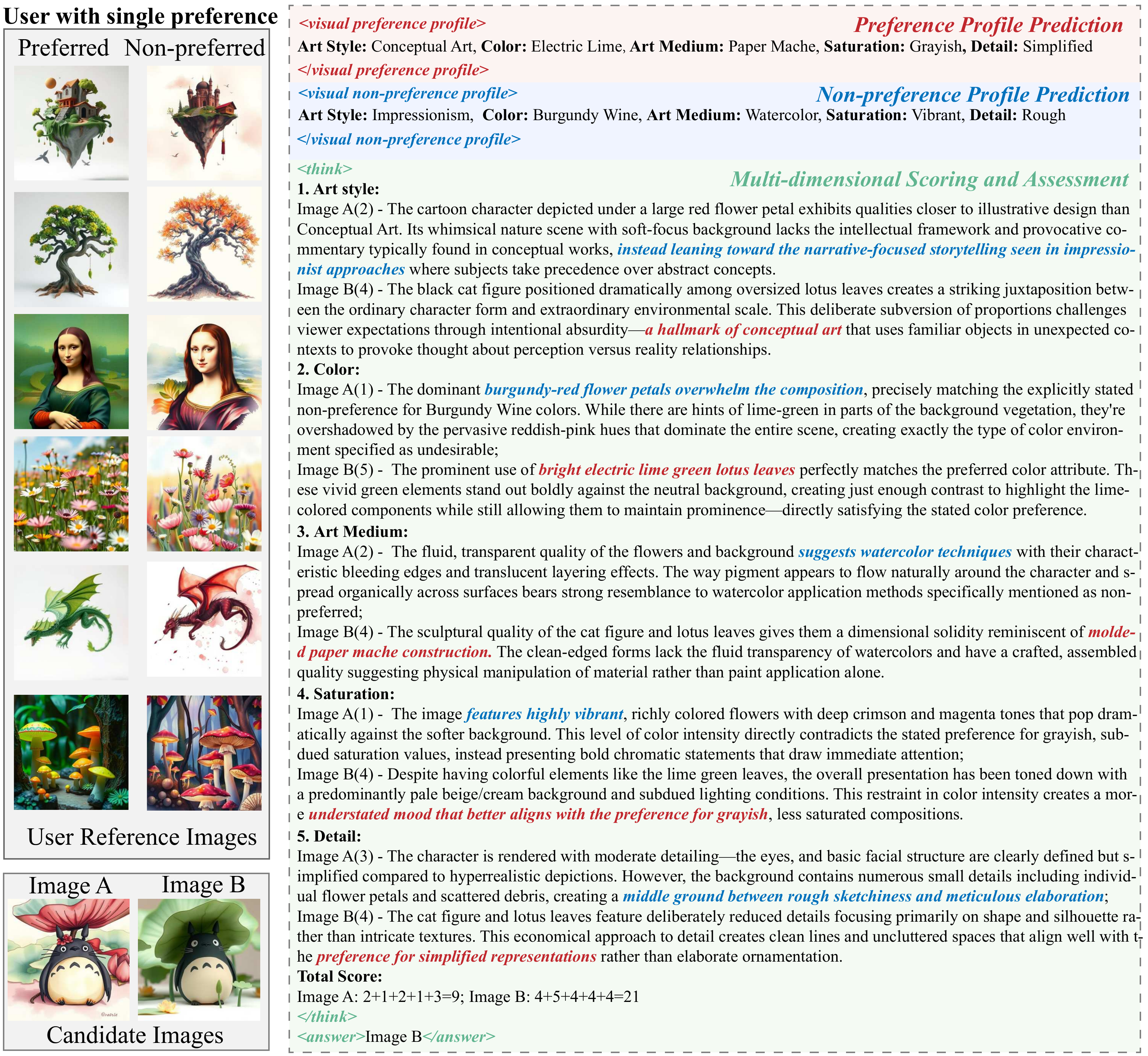}
   \caption{Reasoning-based personalized assessment for users with single preference.}
   \label{supp_sp_case_1}

\end{figure}

\begin{figure}[t]
  \setlength{\belowcaptionskip}{-15pt}
  \captionsetup{skip=2pt} % 仅对当前图片生效
  \centering
   \includegraphics[width=0.95\linewidth]{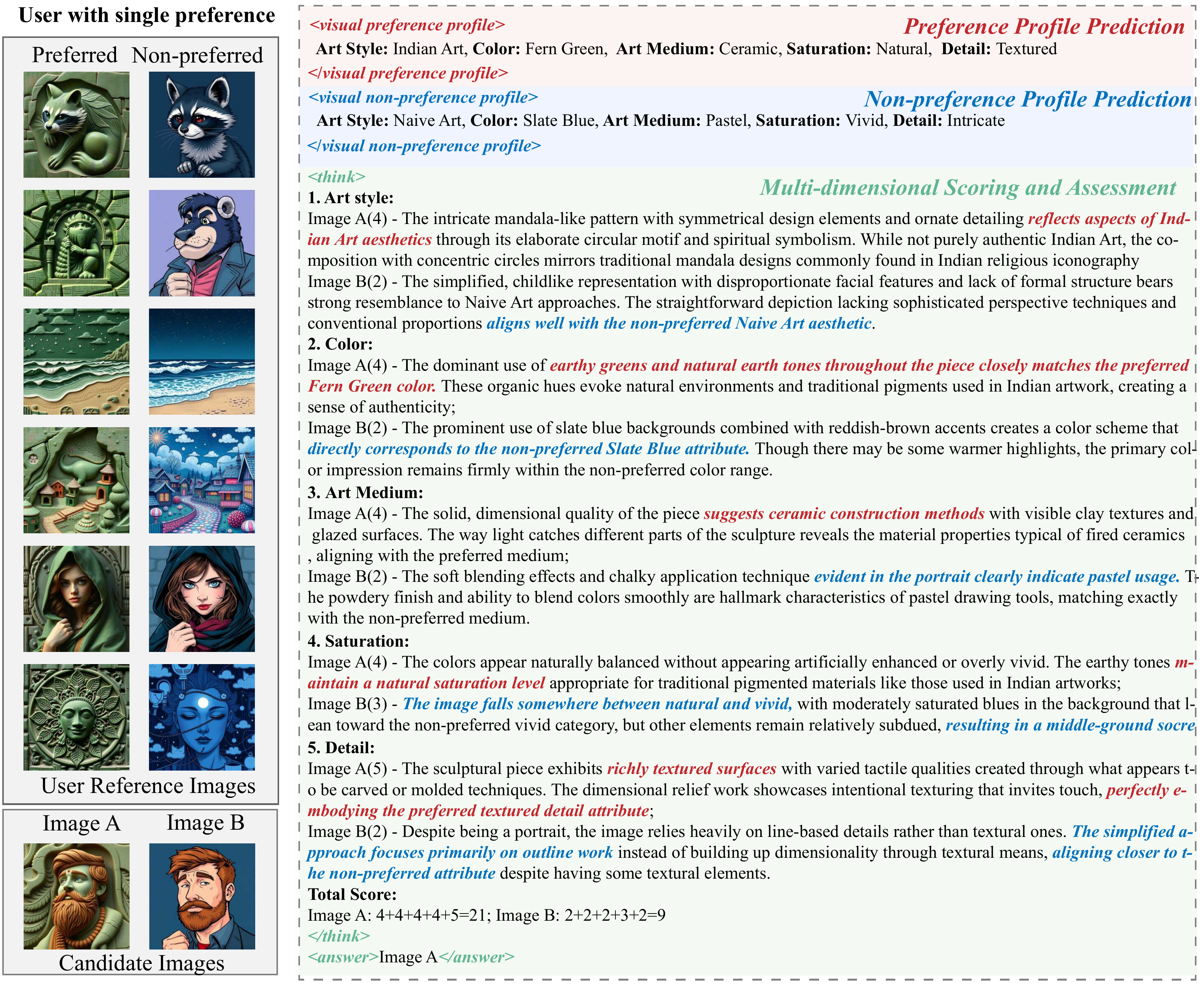}
   \caption{Reasoning-based personalized assessment for users with single preference.}
   \label{supp_sp_case_2}

\end{figure}

\begin{figure}[b]
  \setlength{\belowcaptionskip}{-15pt}
  \captionsetup{skip=2pt} % 仅对当前图片生效
  \centering
   \includegraphics[width=0.95\linewidth]{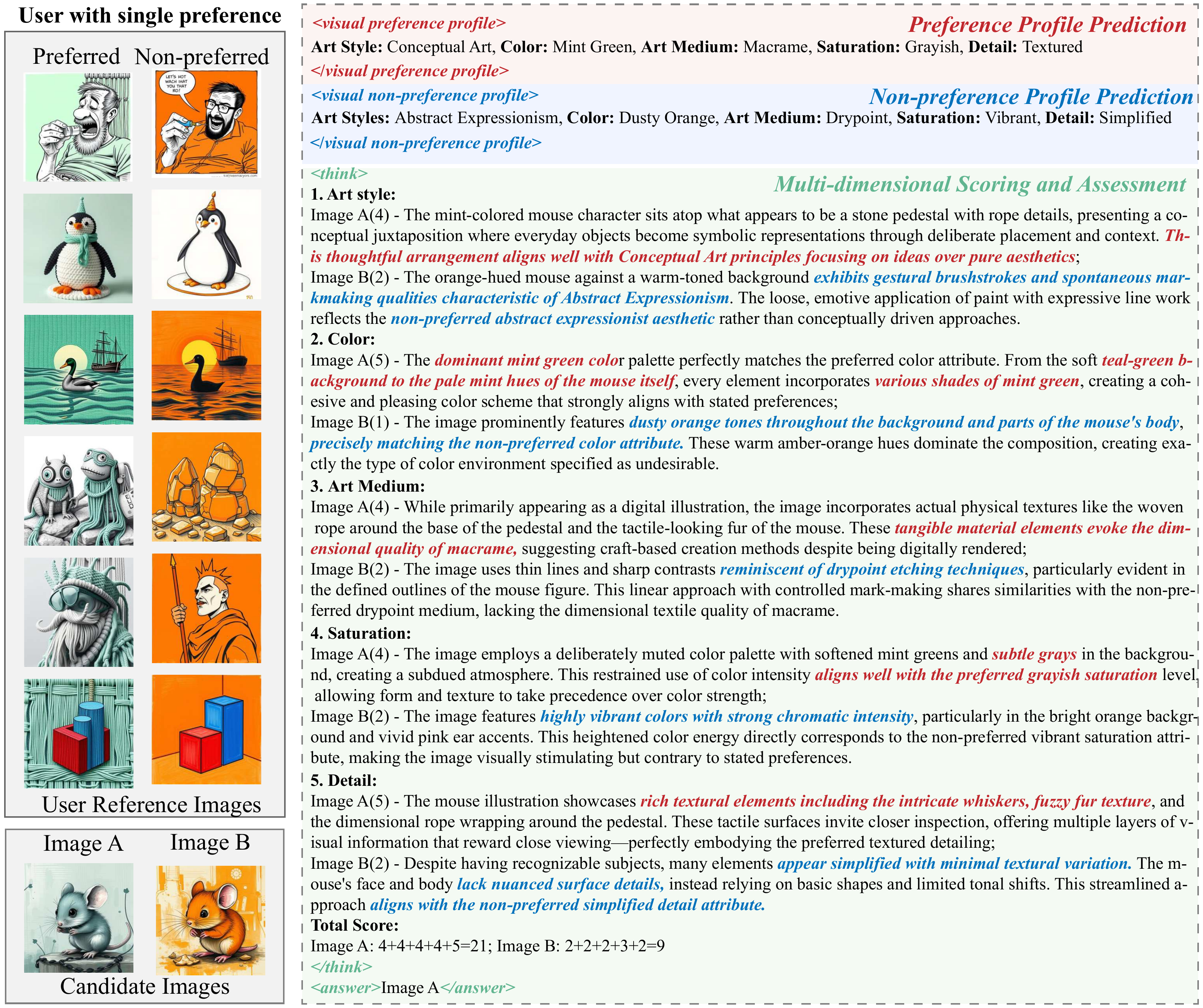}
   \caption{Reasoning-based personalized assessment for users with single preference.}
   \label{supp_sp_case_3}

\end{figure}

We present more visualization results on the PreferImg where users's reference images reflect single preference in \Cref{supp_sp_case_1,supp_sp_case_2,supp_sp_case_3}. It is observed that PreferThinker first predicts the users' preference and non-preference profiles from the reference images, and then provides a reliable assessment and score for candidate images based on these profiles, which not only enables accurate assessment but also provides interpretability.

\subsection{Visualization Results on Users with Multiple Preferences}
\label{sec:supp_visualization_mp}

\begin{figure}[t]
  \setlength{\belowcaptionskip}{-15pt}
  \captionsetup{skip=2pt} % 仅对当前图片生效
  \centering
   \includegraphics[width=0.94\linewidth]{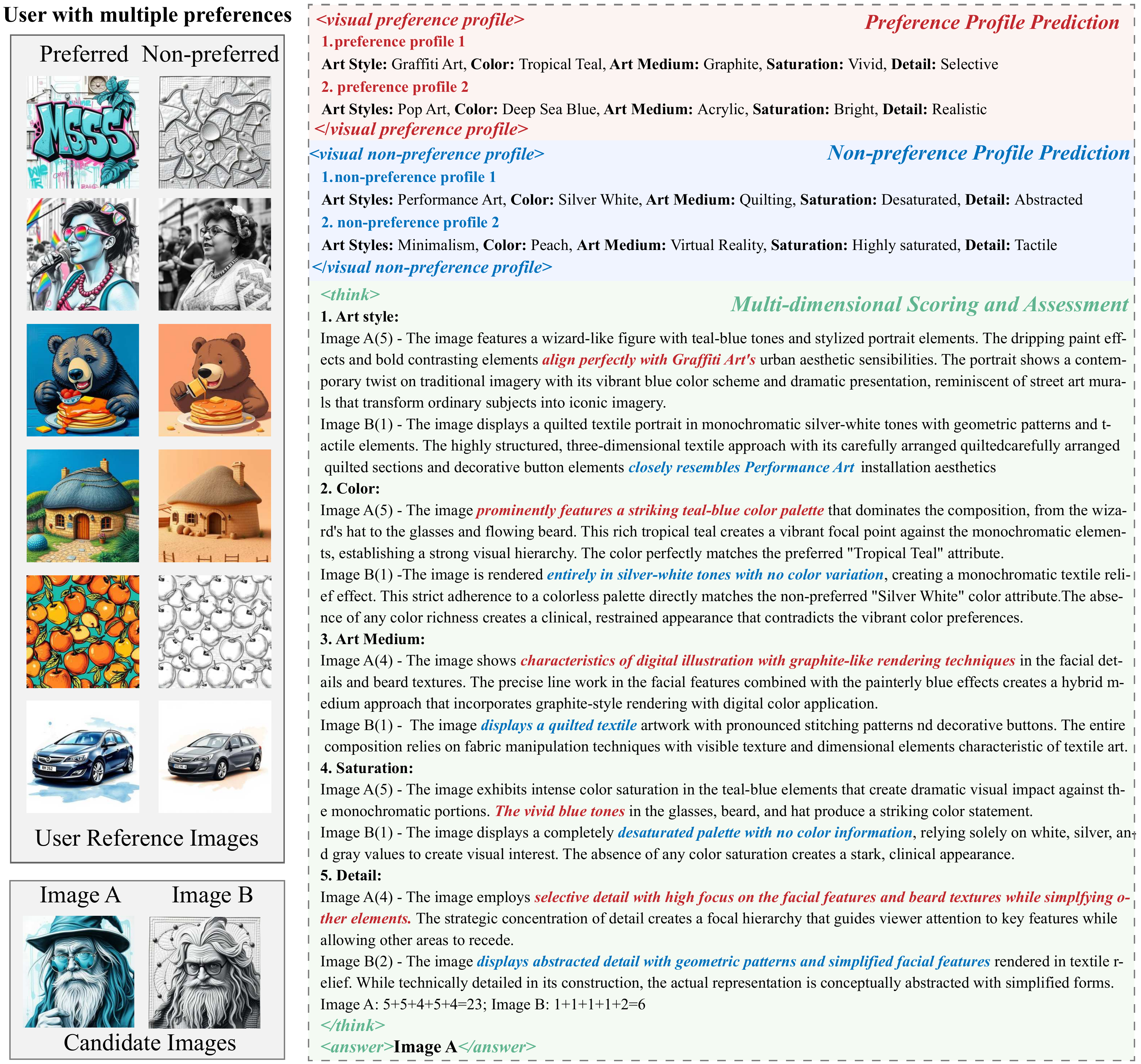}
   \caption{Reasoning-based personalized assessment for users with multiple preferences.}
   \label{supp_mp_case_1}

\end{figure}

\begin{figure}[t]
  \setlength{\belowcaptionskip}{-15pt}
  \captionsetup{skip=2pt} % 仅对当前图片生效
  \centering
   \includegraphics[width=0.94\linewidth]{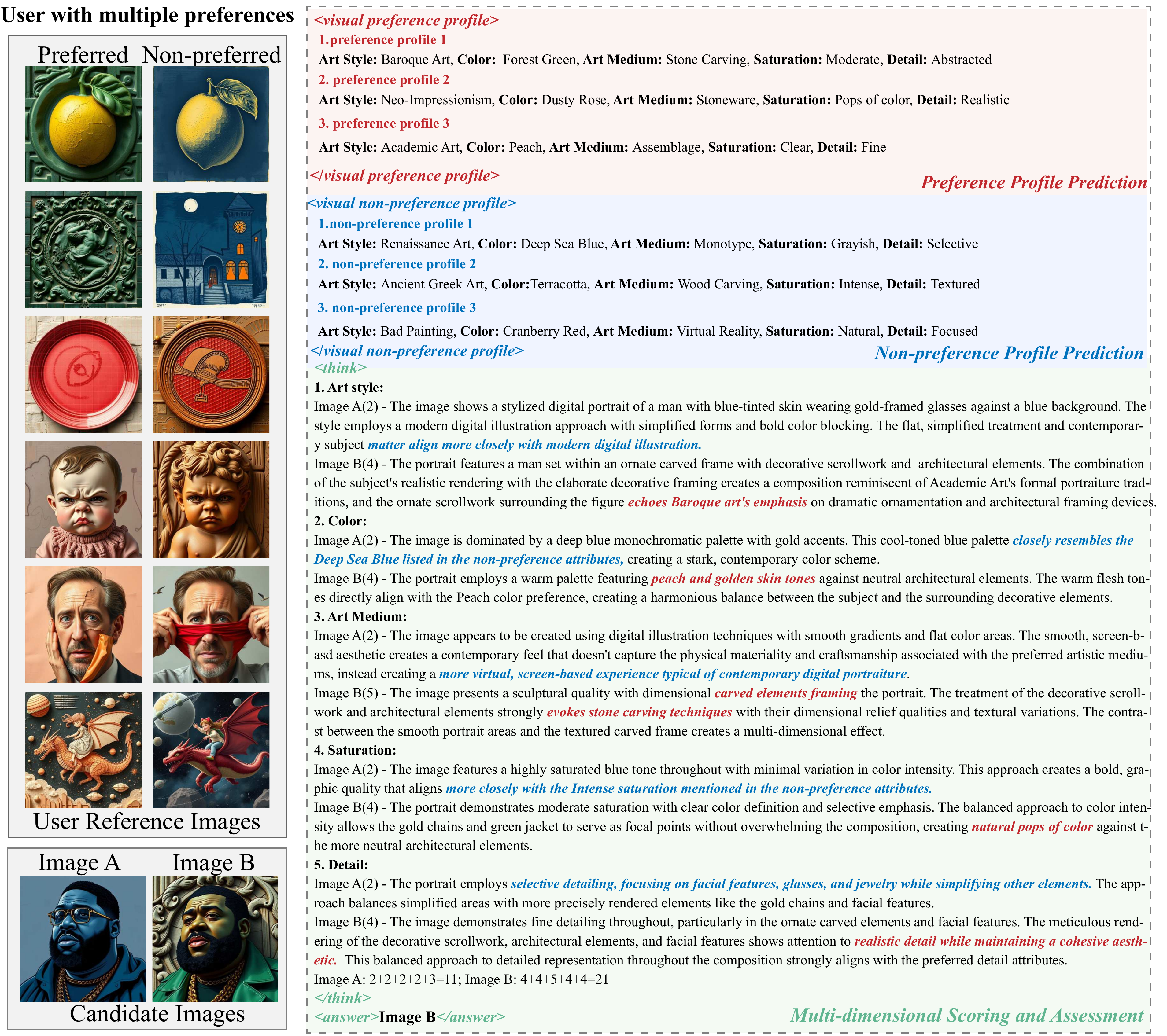}
   \caption{Reasoning-based personalized assessment for users with multiple preferences.}
   \label{supp_mp_case_2}

\end{figure}

We show more visualization results on the PreferImg where users's reference images reflect multiple Preferences in Fig. \ref{supp_mp_case_1} and \ref{supp_mp_case_2}. Our model can accurately predict multiple preference profiles from reference images and provide a reliable assessment and score for candidate images based on these profiles, revealing the effectiveness of PreferThinker for users with multiple preferences.

\subsection{Visualization Results on Real Users}
\label{sec:supp_visualization_real_user}
\begin{figure}[t]

  \setlength{\belowcaptionskip}{-15pt}
  \captionsetup{skip=2pt} % 仅对当前图片生效
  \centering
   \includegraphics[width=0.94\linewidth]{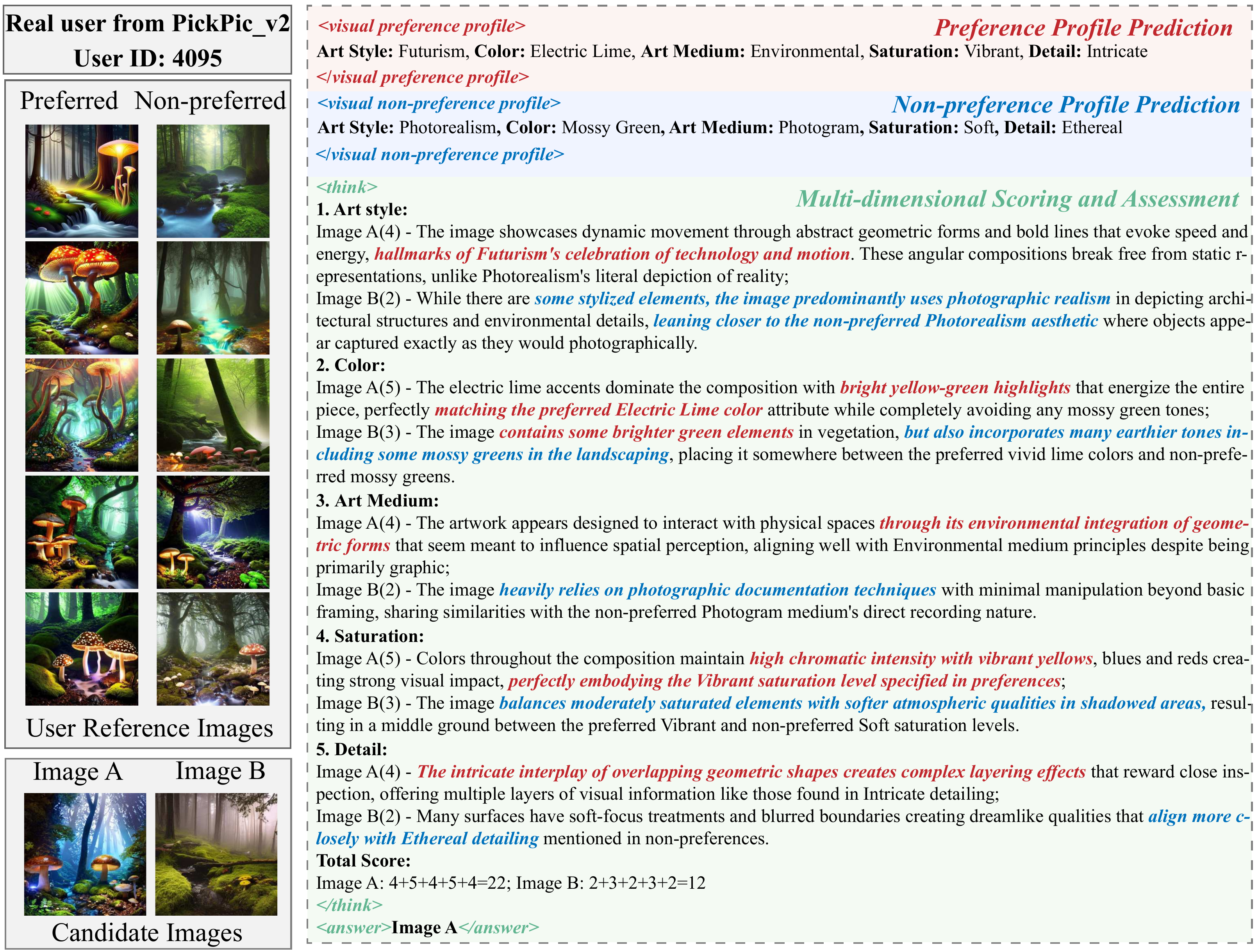}
   \caption{Reasoning-based personalized assessment for Real user from PickaPic.}
   \label{supp_real_user_case_1}
\end{figure}

\begin{figure}[t]
      \vspace{-20pt}
  \setlength{\belowcaptionskip}{-15pt}
  \captionsetup{skip=2pt} % 仅对当前图片生效
  \centering
   \includegraphics[width=0.95\linewidth]{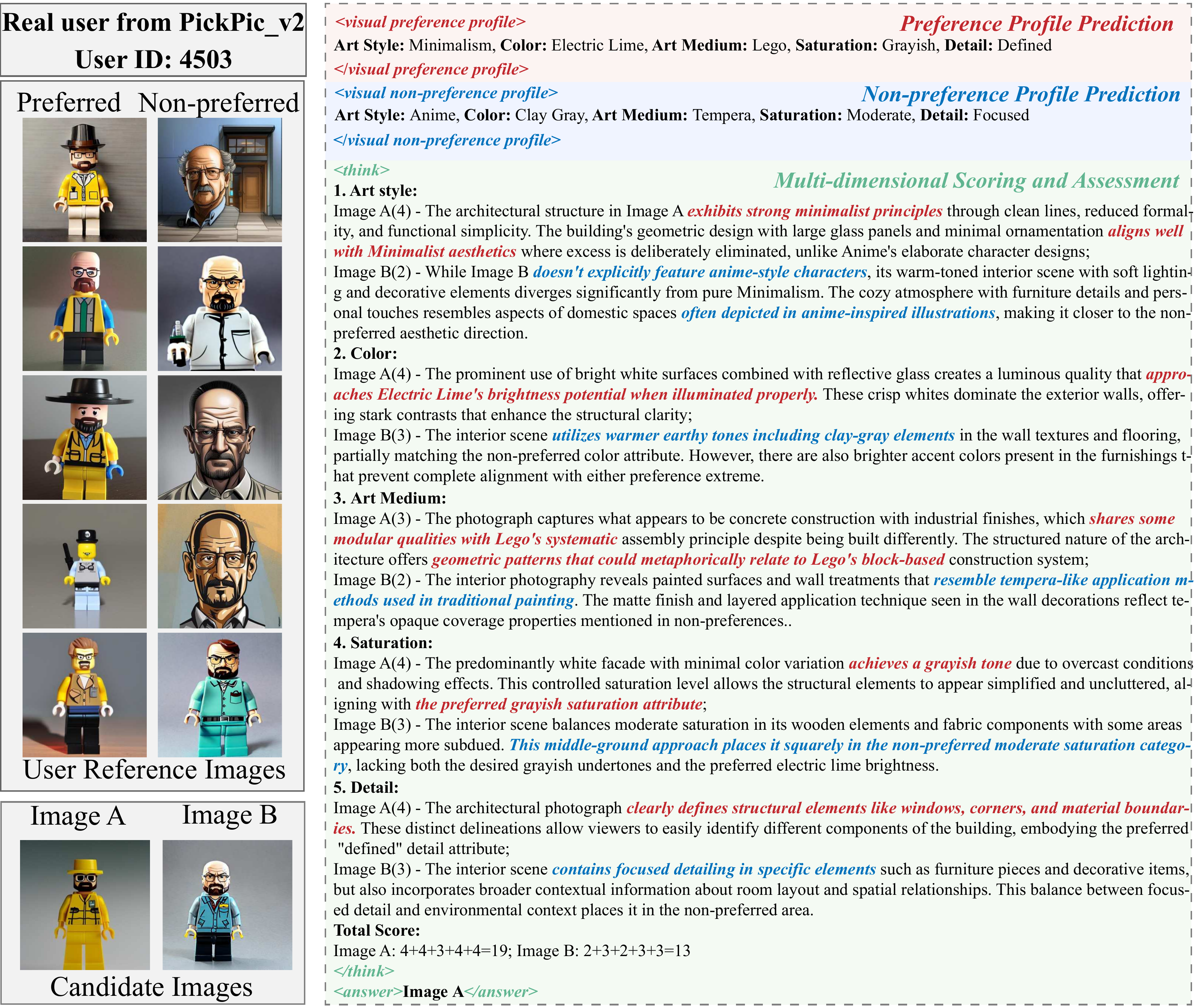}
   \caption{Reasoning-based personalized assessment for Real user from PickaPic.}
   \label{supp_real_user_case_2}

\end{figure}

\begin{figure}[b]
  \setlength{\belowcaptionskip}{-15pt}
  \captionsetup{skip=2pt} % 仅对当前图片生效
  \centering
   \includegraphics[width=0.95\linewidth]{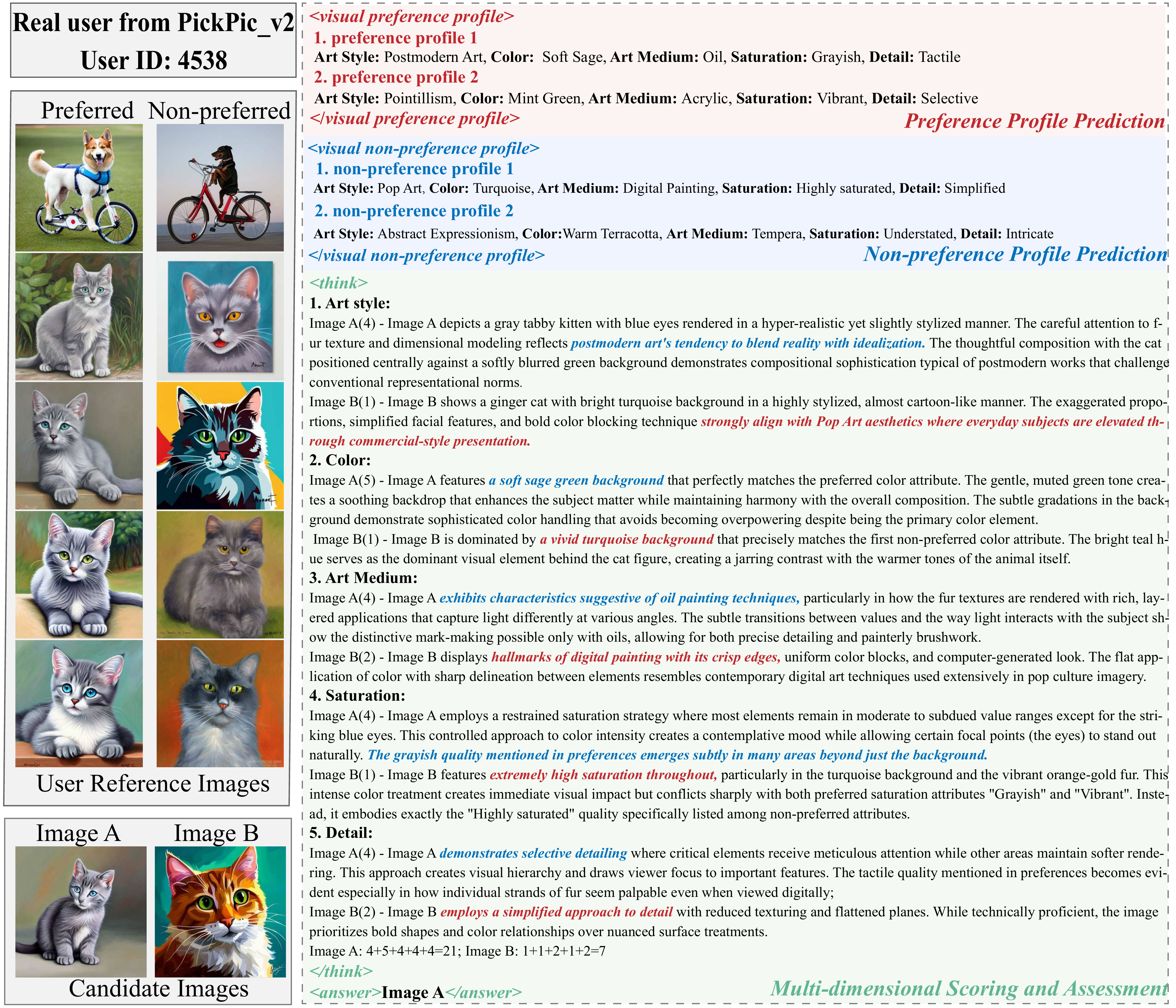}
   \caption{Reasoning-based personalized assessment for Real user from PickaPic.}
   \label{supp_real_user_case_3}

\end{figure}

\begin{figure}[t]
  \setlength{\belowcaptionskip}{-15pt}
  \captionsetup{skip=2pt} % 仅对当前图片生效
  \centering
   \includegraphics[width=0.95\linewidth]{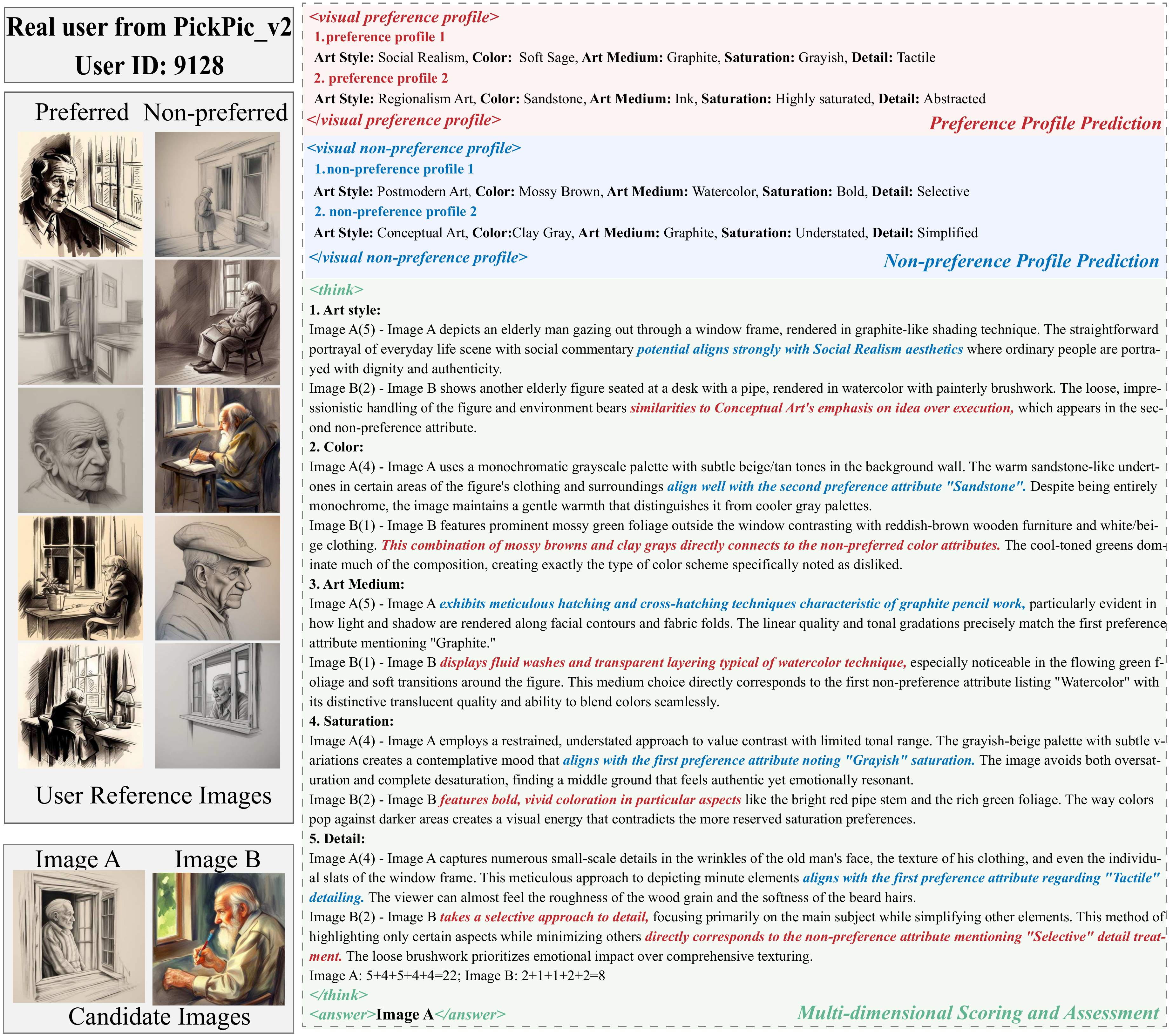}
   \caption{Reasoning-based personalized assessment for Real user from PickaPic.}
   \label{supp_real_user_case_4}

\end{figure}

To demonstrate the generalizability of our method to real-world users, we further utilize the PreferThinker to conduct personalized preference assessment for real users from PickaPic. As shown in \Cref{supp_real_user_case_1,supp_real_user_case_2,supp_real_user_case_3,supp_real_user_case_4} Although the preferences in real users' reference images are more complex, our method can still effectively extract their primary preference profiles and provide a correct assessment.

\section{Discussion}
\label{sec:supp_discussion}

% \subsection{Performance and Speed of PreferThinker without CoT.}
% \label{sec:supp_performance_speed}
% To evaluate the performance and  inference speed of PreferThinker without CoT-style reasoning, we modify the prompt to enable the model to directly assess candidate images based on reference images and output the answer without CoT reasoning. As shown in Table \ref{tab:performance_speed_w/o COt}, Preferthinker maintains strong performance even after removing the CoT-style reasoning, while significantly boosting its inference speed. This is because the model, having been trained with both CoT-based and group-based methods, retains powerful prior knowledge for personalized preference assessment.

% \begin{table}[t]
%     \centering
%       \caption{Performance and inference speed of PreferThinker with and without CoT.}
%           \resizebox{0.5\textwidth}{!}{ % 将宽度设置为与wraptable一致
%     \begin{tabular}{ccc}
% \hline
% \hline
%         Methods & Accuracy (\%) $\uparrow$ & Inference Speed (s) $\downarrow$  \\
%         \hline
% \hline
%         ViPer & 92.4 & 1.63  \\
%         Claude 3.7 & 93.8 & 23.04   \\
%         PreferThinker (w/o CoT)& 94.0 & \textbf{0.75}\\
%         PreferThinker & \textbf{96.6} & 14.51  \\
% \hline
% \hline
%     \end{tabular}
%           }
%     \label{tab:performance_speed_w/o COt}
% \end{table}

\subsection{Importance of Cold Start Initialization.}
\label{sec:supp_cold_start_effectiveness}
\begin{figure}[b]
  \setlength{\belowcaptionskip}{-15pt}
  \captionsetup{skip=2pt} % 仅对当前图片生效
  \centering
   \includegraphics[width=0.95\linewidth]{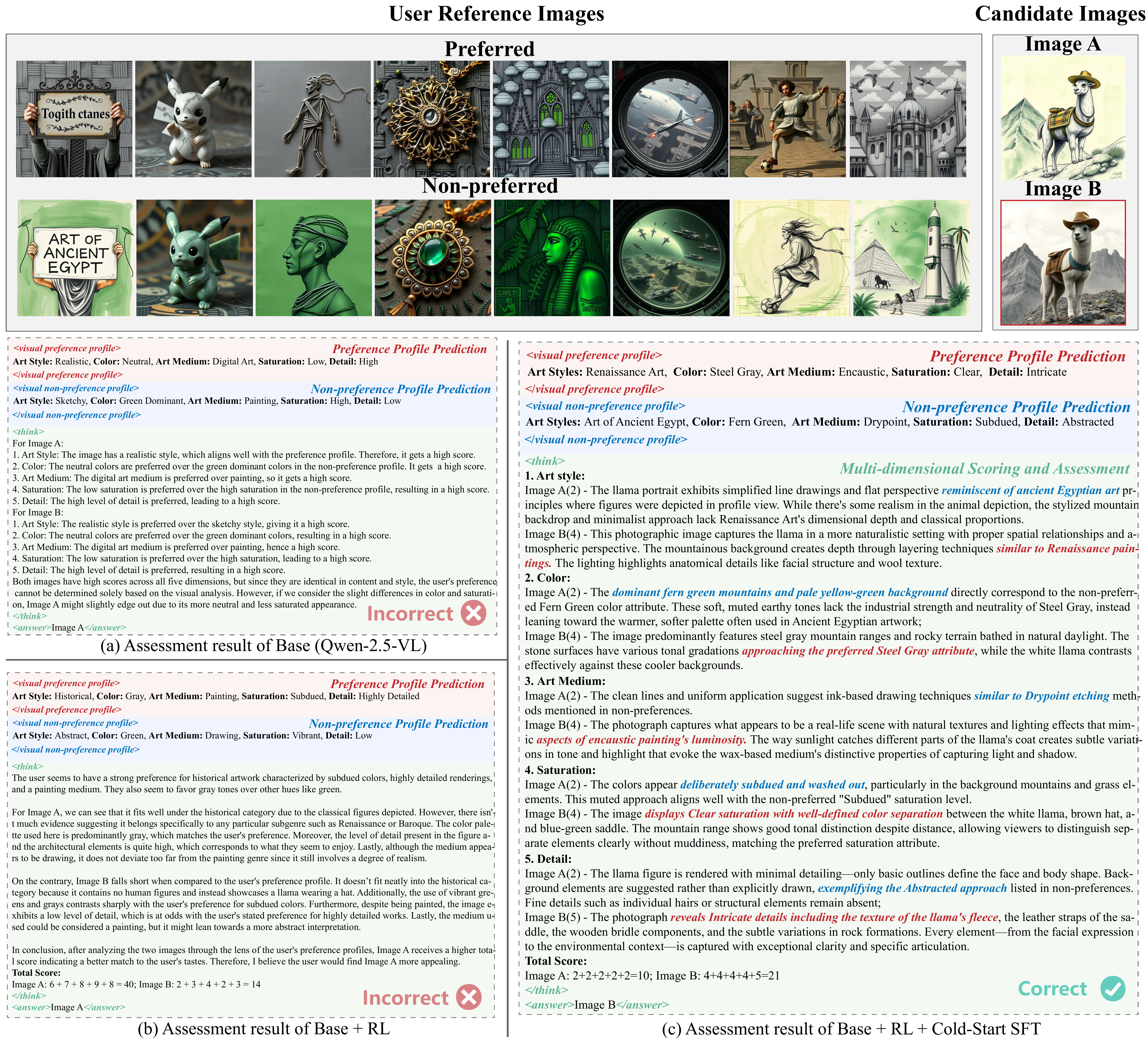}
   \caption{Effectiveness of cold-start SFT for correct and structured reasoning.}
   \label{supp_effectiveness_SFT}

\end{figure}

We further discuss the importance of cold-start SFT for correct structured reasoning. Figure \ref{supp_effectiveness_SFT}(a) shows that directly using the base model (Qwen-2.5-VL) fails to provide numerical scores and reasonable reasoning and leads to incorrect answer. Figure \ref{supp_effectiveness_SFT} (b) shows that while applying reinforcement learning alone on the based model can generate the total numerical scores,  it fails to provide the correct output format and interpretable scores for each visual elements. In contrast, by first teaching the model proper structural reasoning with Supervised Fine-Tuning (SFT) before RL, the model can provide reasonable, explainable scores for each dimension, ultimately leading to a correct final assessment based on the total score, as shown in Figure \ref{supp_effectiveness_SFT} (c). 

\subsection{Transferability to Personalized Assessment and Scoring of Single Image.}
\label{sec:supp_score_single_image}

\begin{figure}[t]
  \setlength{\belowcaptionskip}{-15pt}
  \captionsetup{skip=2pt} % 仅对当前图片生效
  \centering
   \includegraphics[width=0.95\linewidth]{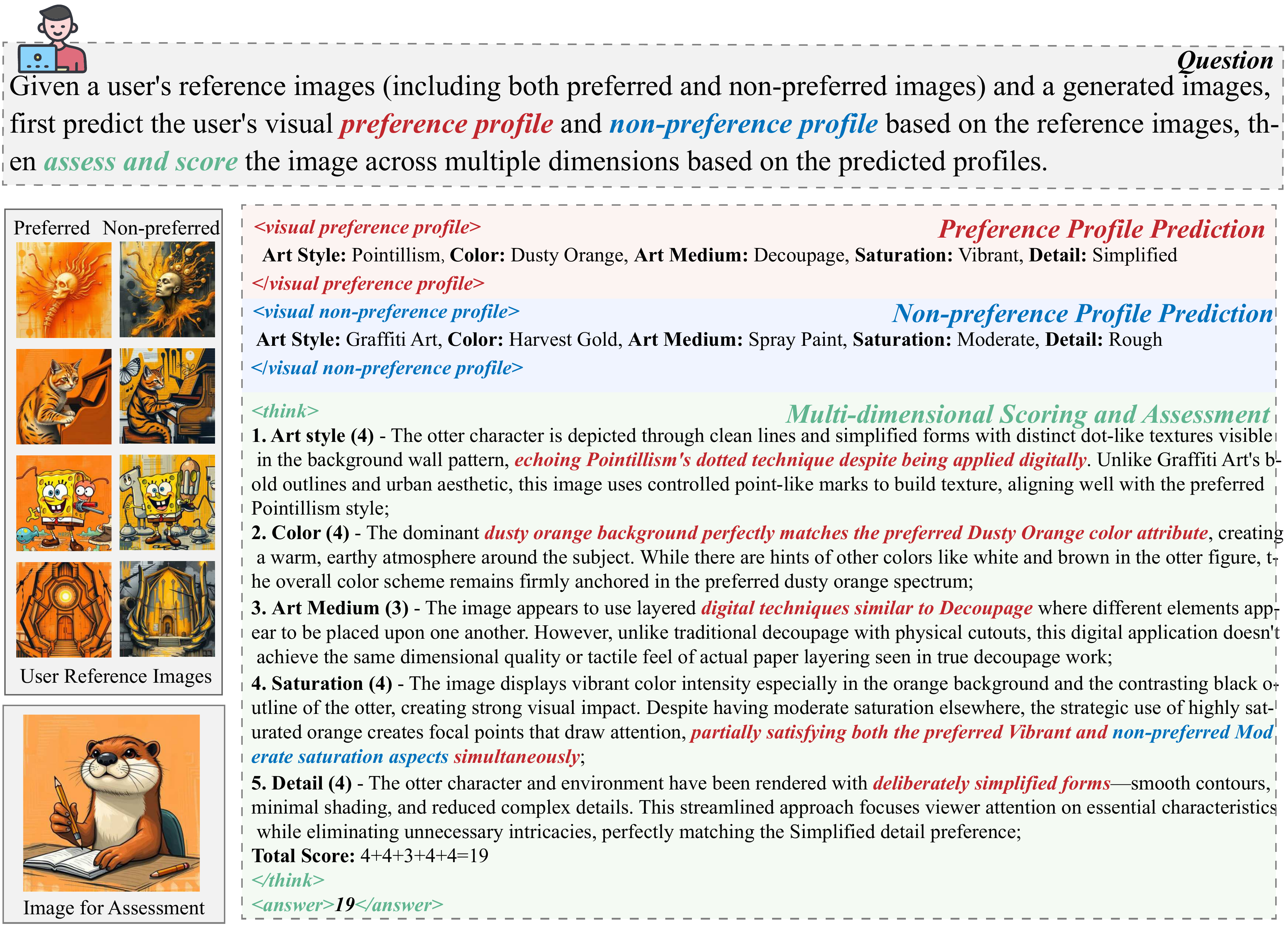}
   \caption{Reasoning-based personalized assessment and scoring for single image.}
   \label{supp_score_case_1}

\end{figure}

\begin{figure}[t]
  \setlength{\belowcaptionskip}{-15pt}
  \captionsetup{skip=2pt} % 仅对当前图片生效
  \centering
   \includegraphics[width=0.95\linewidth]{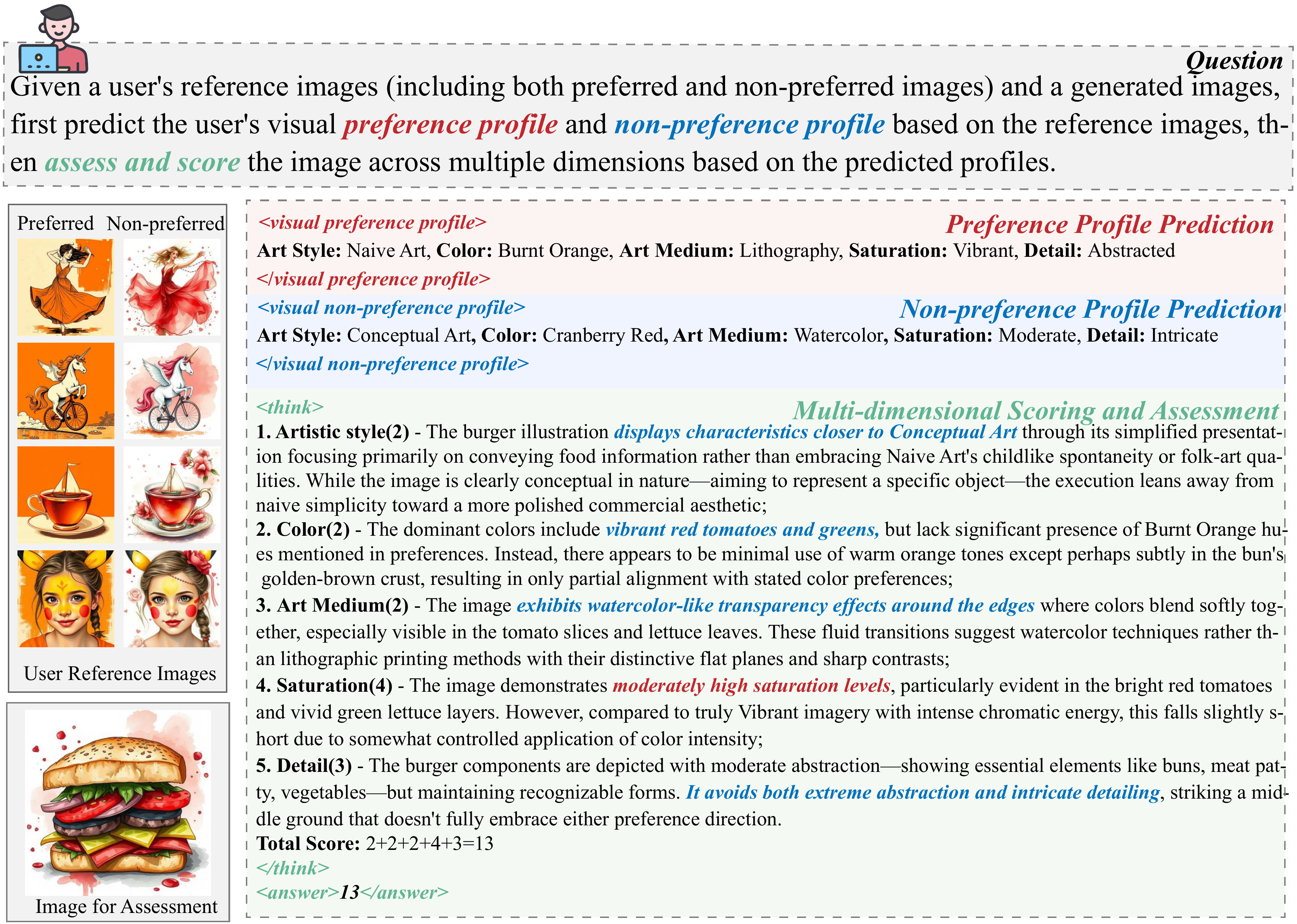}
   \caption{Reasoning-based personalized assessment and scoring for single image.}
   \label{supp_score_case_2}

\end{figure}

We further study the transferability of PreferThinker towards personalized assessment and scoring of single image. We directly modify the question prompt to enable the PreferThinker to provide an assessment and score for a single generated image. The visualization results are shown in Fig. \ref{supp_score_case_1} and Fig. \ref{supp_score_case_2}. We can observe that PreferThinker can flexibly provide an interpretable assessment and score for a single image.

\subsection{Personalized Image Generation with Predicted Profile.}
\label{sec:supp_personalized_image_generation}
\begin{figure}[t]
  \setlength{\belowcaptionskip}{-15pt}
  \captionsetup{skip=2pt} % 仅对当前图片生效
  \centering
   \includegraphics[width=0.95\linewidth]{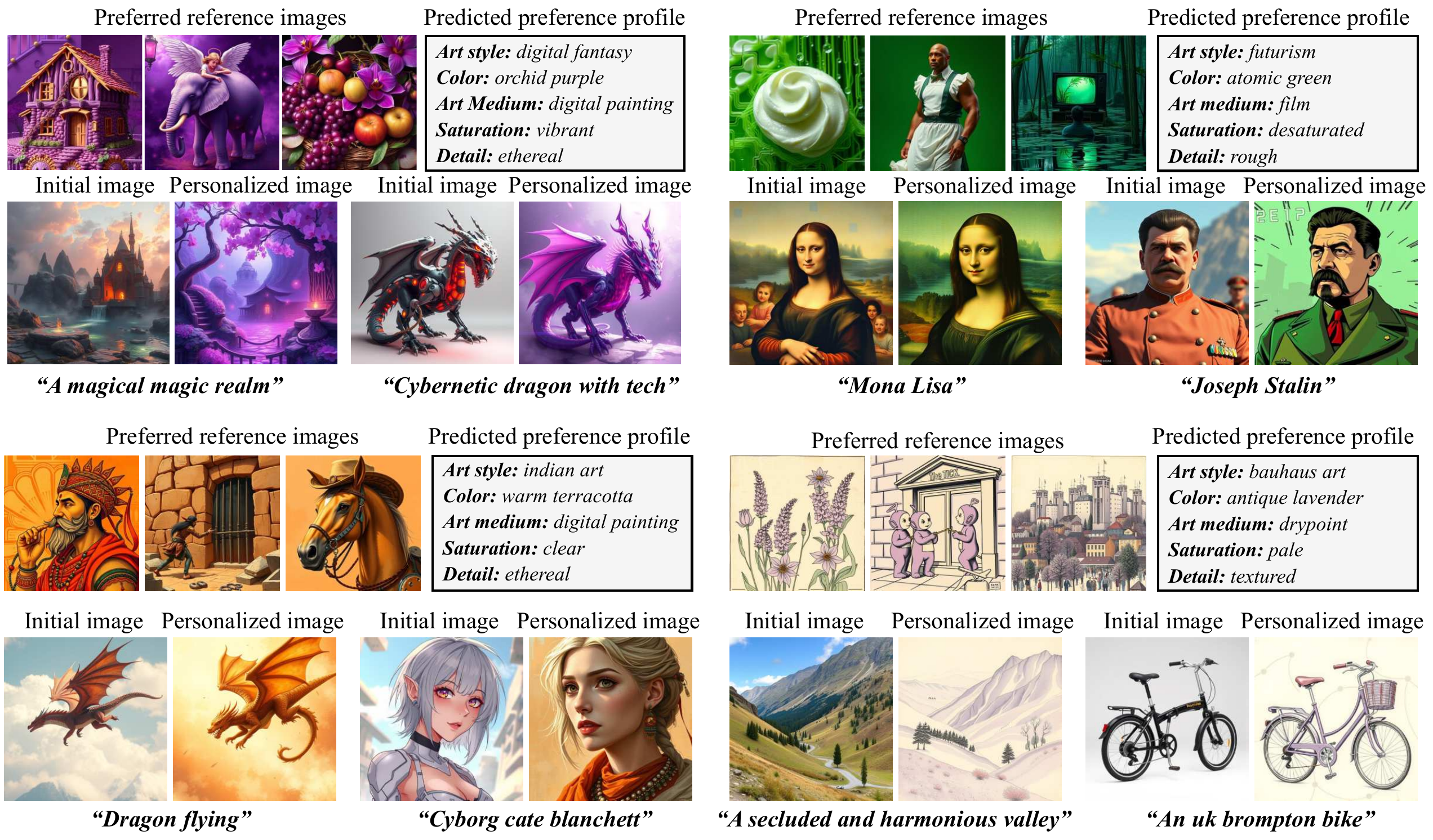}
   \caption{Personalized image generalization with the predicted profile.}
   \label{supp_personalized_generation}

\end{figure}

To further demonstrate the benefits of the proposed preference profile, we employ the predicted preference profile to achieve personalized image generation in Fig. \ref{supp_personalized_generation}. We first predict the user's preference profile based on reference images. Then, we optimize the initial text based on the profile to generate  personalized images.The results show that the personalized images align well with reference images, revealing the potential of the proposed method to facilitate personalized generation. 
\subsection{Limitation and Future work}
\label{sec:Limitation_and_Future_work}
\begin{figure}[t]
  \setlength{\belowcaptionskip}{-15pt}
  \captionsetup{skip=2pt} % 仅对当前图片生效
  \centering
   \includegraphics[width=0.95\linewidth]{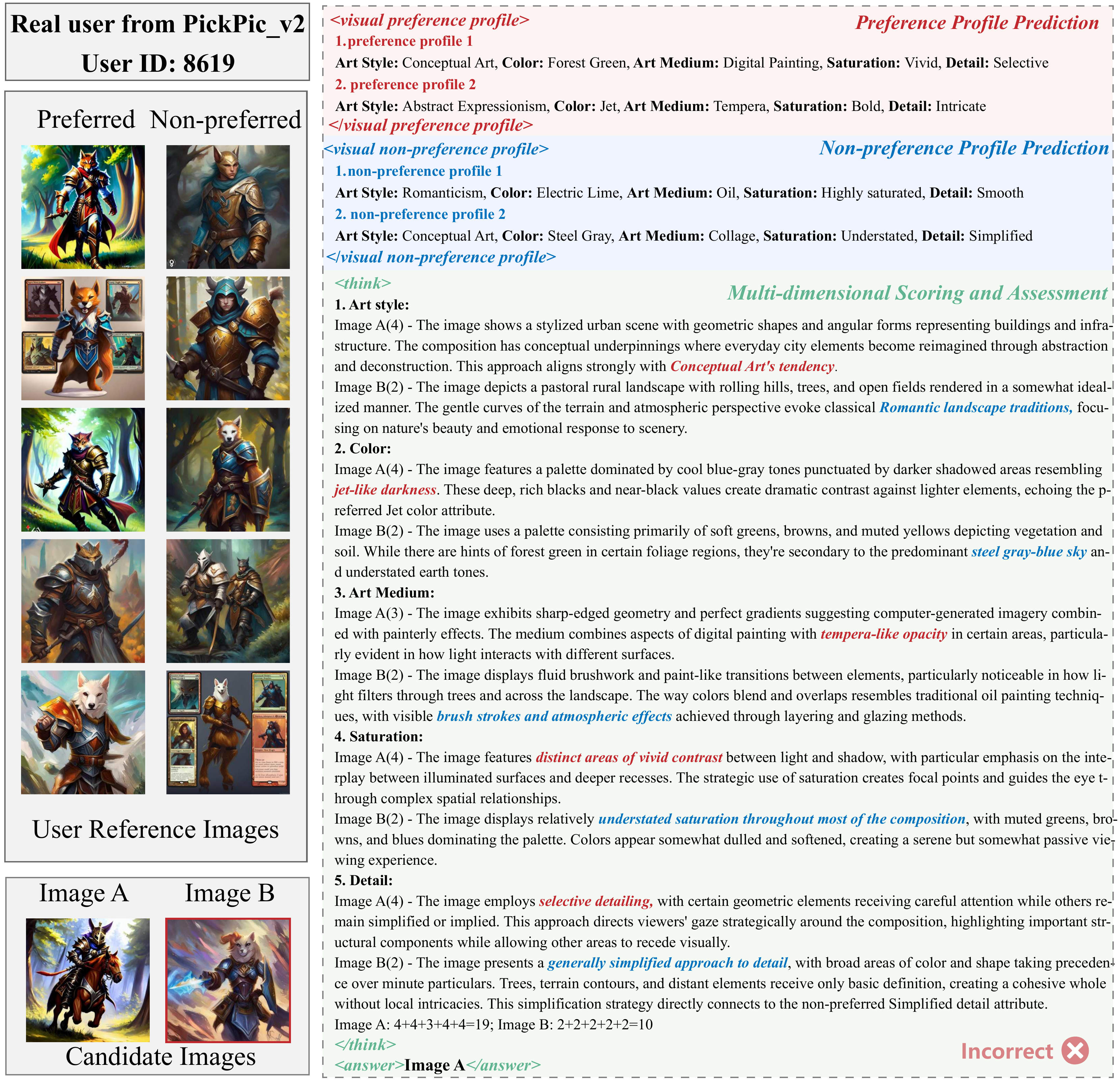}
   \caption{Failed case of the Preferthinker for real users with complex preferences.}
   \label{limitation}

\end{figure}

The main limitation of our method is that the personalized preference dataset built on synthetic profiles may not fully capture the complex and subtle real human preferences. Our current preference profile only contains five key visual attributes that influence human visual preferences. However, real users' preferences are often complex and subtle, shaped by multiple factors. Beyond visual attributes, they are also closely related to image content semantics and users’ backgrounds (e.g., age, gender and culture). Consequently, our method may struggle with complex and hard-to-articulate real user preferences, as shown in Fig. \ref{limitation}. In future work, we will incorporate additional key factors into preference profiles to more comprehensively capture real users' preferences.

\section{Use of LLMs}
   \label{supp_use_llm}

We only used LLMs to polish and correct the text and grammar in the Abstract and Introduction sections.


%% file: iclr2026_conference.bib
@article{heusel2017gans,
  title={Gans trained by a two time-scale update rule converge to a local nash equilibrium},
  author={Heusel, Martin and Ramsauer, Hubert and Unterthiner, Thomas and Nessler, Bernhard and Hochreiter, Sepp},
  journal={Advances in neural information processing systems},
  volume={30},
  year={2017}
}

@inproceedings{radford2021learning,
  title={Learning transferable visual models from natural language supervision},
  author={Radford, Alec and Kim, Jong Wook and Hallacy, Chris and Ramesh, Aditya and Goh, Gabriel and Agarwal, Sandhini and Sastry, Girish and Askell, Amanda and Mishkin, Pamela and Clark, Jack and others},
  booktitle={International conference on machine learning},
  pages={8748--8763},
  year={2021},
  organization={PmLR}
}

@article{xu2024visionreward,
  title={Visionreward: Fine-grained multi-dimensional human preference learning for image and video generation},
  author={Xu, Jiazheng and Huang, Yu and Cheng, Jiale and Yang, Yuanming and Xu, Jiajun and Wang, Yuan and Duan, Wenbo and Yang, Shen and Jin, Qunlin and Li, Shurun and others},
  journal={arXiv preprint arXiv:2412.21059},
  year={2024}
}

@article{wang2025unified,
  title={Unified reward model for multimodal understanding and generation},
  author={Wang, Yibin and Zang, Yuhang and Li, Hao and Jin, Cheng and Wang, Jiaqi},
  journal={arXiv preprint arXiv:2503.05236},
  year={2025}
}

@article{zhou2025multimodal,
  title={Multimodal LLMs as Customized Reward Models for Text-to-Image Generation},
  author={Zhou, Shijie and Zhang, Ruiyi and Zhu, Huaisheng and Kveton, Branislav and Zhou, Yufan and Gu, Jiuxiang and Chen, Jian and Chen, Changyou},
  journal={arXiv preprint arXiv:2507.21391},
  year={2025}
}

@article{wang2025unifiedthink,
  title={Unified multimodal chain-of-thought reward model through reinforcement fine-tuning},
  author={Wang, Yibin and Li, Zhimin and Zang, Yuhang and Wang, Chunyu and Lu, Qinglin and Jin, Cheng and Wang, Jiaqi},
  journal={arXiv preprint arXiv:2505.03318},
  year={2025}
}

@article{gambashidze2025listener,
  title={Listener-Rewarded Thinking in VLMs for Image Preferences},
  author={Gambashidze, Alexander and Pengyi, Li and Skripkin, Matvey and Galichin, Andrey and Gusarov, Anton and Sobolev, Konstantin and Kuznetsov, Andrey and Oseledets, Ivan},
  journal={arXiv preprint arXiv:2506.22832},
  year={2025}
}

@inproceedings{salehi2024viper,
  title={Viper: Visual personalization of generative models via individual preference learning},
  author={Salehi, Sogand and Shafiei, Mahdi and Yeo, Teresa and Bachmann, Roman and Zamir, Amir},
  booktitle={European Conference on Computer Vision},
  pages={391--406},
  year={2024},
  organization={Springer}
}

@article{xu2023imagereward,
  title={Imagereward: Learning and evaluating human preferences for text-to-image generation},
  author={Xu, Jiazheng and Liu, Xiao and Wu, Yuchen and Tong, Yuxuan and Li, Qinkai and Ding, Ming and Tang, Jie and Dong, Yuxiao},
  journal={Advances in Neural Information Processing Systems},
  volume={36},
  pages={15903--15935},
  year={2023}
}

@article{kirstain2023pick,
  title={Pick-a-pic: An open dataset of user preferences for text-to-image generation},
  author={Kirstain, Yuval and Polyak, Adam and Singer, Uriel and Matiana, Shahbuland and Penna, Joe and Levy, Omer},
  journal={Advances in neural information processing systems},
  volume={36},
  pages={36652--36663},
  year={2023}
}

@inproceedings{wu2023human,
  title={Human preference score: Better aligning text-to-image models with human preference},
  author={Wu, Xiaoshi and Sun, Keqiang and Zhu, Feng and Zhao, Rui and Li, Hongsheng},
  booktitle={Proceedings of the IEEE/CVF International Conference on Computer Vision},
  pages={2096--2105},
  year={2023}
}

@article{wu2023human2,
  title={Human preference score v2: A solid benchmark for evaluating human preferences of text-to-image synthesis},
  author={Wu, Xiaoshi and Hao, Yiming and Sun, Keqiang and Chen, Yixiong and Zhu, Feng and Zhao, Rui and Li, Hongsheng},
  journal={arXiv preprint arXiv:2306.09341},
  year={2023}
}

@inproceedings{li2024stable,
  title={Stable Preference: Redefining Training Paradigm of Human Preference Model for Text-to-Image Synthesis},
  author={Li, Hanting and Niu, Hongjing and Zhao, Feng},
  booktitle={European Conference on Computer Vision},
  pages={250--266},
  year={2024},
  organization={Springer}
}

@article{bahng2025cycle,
  title={Cycle Consistency as Reward: Learning Image-Text Alignment without Human Preferences},
  author={Bahng, Hyojin and Chan, Caroline and Durand, Fredo and Isola, Phillip},
  journal={arXiv preprint arXiv:2506.02095},
  year={2025}
}

@article{ba2025enhancing,
  title={Enhancing Reward Models for High-quality Image Generation: Beyond Text-Image Alignment},
  author={Ba, Ying and Zhang, Tianyu and Bai, Yalong and Mo, Wenyi and Liang, Tao and Su, Bing and Wen, Ji-Rong},
  journal={arXiv preprint arXiv:2507.19002},
  year={2025}
}

@inproceedings{zhang2024learning,
  title={Learning multi-dimensional human preference for text-to-image generation},
  author={Zhang, Sixian and Wang, Bohan and Wu, Junqiang and Li, Yan and Gao, Tingting and Zhang, Di and Wang, Zhongyuan},
  booktitle={Proceedings of the IEEE/CVF Conference on Computer Vision and Pattern Recognition},
  pages={8018--8027},
  year={2024}
}

@inproceedings{liang2024rich,
  title={Rich human feedback for text-to-image generation},
  author={Liang, Youwei and He, Junfeng and Li, Gang and Li, Peizhao and Klimovskiy, Arseniy and Carolan, Nicholas and Sun, Jiao and Pont-Tuset, Jordi and Young, Sarah and Yang, Feng and others},
  booktitle={Proceedings of the IEEE/CVF Conference on Computer Vision and Pattern Recognition},
  pages={19401--19411},
  year={2024}
}

@article{li2025q,
  title={Q-insight: Understanding image quality via visual reinforcement learning},
  author={Li, Weiqi and Zhang, Xuanyu and Zhao, Shijie and Zhang, Yabin and Li, Junlin and Zhang, Li and Zhang, Jian},
  journal={arXiv preprint arXiv:2503.22679},
  year={2025}
}

@article{wu2025visualquality,
  title={VisualQuality-R1: Reasoning-Induced Image Quality Assessment via Reinforcement Learning to Rank},
  author={Wu, Tianhe and Zou, Jian and Liang, Jie and Zhang, Lei and Ma, Kede},
  journal={arXiv preprint arXiv:2505.14460},
  year={2025}
}

@article{cai2025q,
  title={Q-Ponder: A Unified Training Pipeline for Reasoning-based Visual Quality Assessment},
  author={Cai, Zhuoxuan and Zhang, Jian and Yuan, Xinbin and Jiang, Peng-Tao and Chen, Wenxiang and Tang, Bowen and Yao, Lujian and Wang, Qiyuan and Chen, Jinwen and Li, Bo},
  journal={arXiv preprint arXiv:2506.05384},
  year={2025}
}

@article{jiang2025rex,
  title={Rex-Thinker: Grounded Object Referring via Chain-of-Thought Reasoning},
  author={Jiang, Qing and Chen, Xingyu and Zeng, Zhaoyang and Yu, Junzhi and Zhang, Lei},
  journal={arXiv preprint arXiv:2506.04034},
  year={2025}
}

@article{shen2025vlm,
  title={Vlm-r1: A stable and generalizable r1-style large vision-language model},
  author={Shen, Haozhan and Liu, Peng and Li, Jingcheng and Fang, Chunxin and Ma, Yibo and Liao, Jiajia and Shen, Qiaoli and Zhang, Zilun and Zhao, Kangjia and Zhang, Qianqian and others},
  journal={arXiv preprint arXiv:2504.07615},
  year={2025}
}

@article{zhang2025improving,
  title={Improving the Reasoning of Multi-Image Grounding in MLLMs via Reinforcement Learning},
  author={Zhang, Bob and Li, Haoran and Zhang, Tao and Yan, Cilin and Cai, Jiayin and Jiang, Xiaolong and Hao, Yanbin},
  journal={arXiv preprint arXiv:2507.00748},
  year={2025}
}

@article{huang2025vision,
  title={Vision-r1: Incentivizing reasoning capability in multimodal large language models},
  author={Huang, Wenxuan and Jia, Bohan and Zhai, Zijie and Cao, Shaosheng and Ye, Zheyu and Zhao, Fei and Xu, Zhe and Hu, Yao and Lin, Shaohui},
  journal={arXiv preprint arXiv:2503.06749},
  year={2025}
}

@article{liu2025visual,
  title={Visual-rft: Visual reinforcement fine-tuning},
  author={Liu, Ziyu and Sun, Zeyi and Zang, Yuhang and Dong, Xiaoyi and Cao, Yuhang and Duan, Haodong and Lin, Dahua and Wang, Jiaqi},
  journal={arXiv preprint arXiv:2503.01785},
  year={2025}
}

@article{yu2025perception,
  title={Perception-r1: Pioneering perception policy with reinforcement learning},
  author={Yu, En and Lin, Kangheng and Zhao, Liang and Yin, Jisheng and Wei, Yana and Peng, Yuang and Wei, Haoran and Sun, Jianjian and Han, Chunrui and Ge, Zheng and others},
  journal={arXiv preprint arXiv:2504.07954},
  year={2025}
}

@article{liu2025seg,
  title={Seg-zero: Reasoning-chain guided segmentation via cognitive reinforcement},
  author={Liu, Yuqi and Peng, Bohao and Zhong, Zhisheng and Yue, Zihao and Lu, Fanbin and Yu, Bei and Jia, Jiaya},
  journal={arXiv preprint arXiv:2503.06520},
  year={2025}
}

@article{you2025seg,
  title={Seg-R1: Segmentation Can Be Surprisingly Simple with Reinforcement Learning},
  author={You, Zuyao and Wu, Zuxuan},
  journal={arXiv preprint arXiv:2506.22624},
  year={2025}
}

@article{chen2025chart,
  title={Chart-R1: Chain-of-Thought Supervision and Reinforcement for Advanced Chart Reasoner},
  author={Chen, Lei and Zhao, Xuanle and Zeng, Zhixiong and Huang, Jing and Zhong, Yufeng and Ma, Lin},
  journal={arXiv preprint arXiv:2507.15509},
  year={2025}
}

@article{pan2025medvlm,
  title={Medvlm-r1: Incentivizing medical reasoning capability of vision-language models (vlms) via reinforcement learning},
  author={Pan, Jiazhen and Liu, Che and Wu, Junde and Liu, Fenglin and Zhu, Jiayuan and Li, Hongwei Bran and Chen, Chen and Ouyang, Cheng and Rueckert, Daniel},
  journal={arXiv preprint arXiv:2502.19634},
  year={2025}
}

@article{ni2025point,
  title={Point-rft: Improving multimodal reasoning with visually grounded reinforcement finetuning},
  author={Ni, Minheng and Yang, Zhengyuan and Li, Linjie and Lin, Chung-Ching and Lin, Kevin and Zuo, Wangmeng and Wang, Lijuan},
  journal={arXiv preprint arXiv:2505.19702},
  year={2025}
}

@article{guo2025decoupled,
  title={Decoupled Visual Interpretation and Linguistic Reasoning for Math Problem Solving},
  author={Guo, Zixian and Liu, Ming and Ji, Zhilong and Bai, Jinfeng and Zhang, Lei and Zuo, Wangmeng},
  journal={arXiv preprint arXiv:2505.17609},
  year={2025}
}

@article{li2025vision,
  title={Vision Matters: Simple Visual Perturbations Can Boost Multimodal Math Reasoning},
  author={Li, Yuting and Wei, Lai and Zheng, Kaipeng and Huang, Jingyuan and Kong, Linghe and Sun, Lichao and Huang, Weiran},
  journal={arXiv preprint arXiv:2506.09736},
  year={2025}
}

@article{zhang2025r1,
  title={R1-vl: Learning to reason with multimodal large language models via step-wise group relative policy optimization},
  author={Zhang, Jingyi and Huang, Jiaxing and Yao, Huanjin and Liu, Shunyu and Zhang, Xikun and Lu, Shijian and Tao, Dacheng},
  journal={arXiv preprint arXiv:2503.12937},
  year={2025}
}

@article{guo2025deepseek,
  title={Deepseek-r1: Incentivizing reasoning capability in llms via reinforcement learning},
  author={Guo, Daya and Yang, Dejian and Zhang, Haowei and Song, Junxiao and Zhang, Ruoyu and Xu, Runxin and Zhu, Qihao and Ma, Shirong and Wang, Peiyi and Bi, Xiao and others},
  journal={arXiv preprint arXiv:2501.12948},
  year={2025}
}

@article{shao2024deepseekmath,
  title={Deepseekmath: Pushing the limits of mathematical reasoning in open language models},
  author={Shao, Zhihong and Wang, Peiyi and Zhu, Qihao and Xu, Runxin and Song, Junxiao and Bi, Xiao and Zhang, Haowei and Zhang, Mingchuan and Li, YK and Wu, Yang and others},
  journal={arXiv preprint arXiv:2402.03300},
  year={2024}
}

@article{schulman2017proximal,
  title={Proximal policy optimization algorithms},
  author={Schulman, John and Wolski, Filip and Dhariwal, Prafulla and Radford, Alec and Klimov, Oleg},
  journal={arXiv preprint arXiv:1707.06347},
  year={2017}
}

@article{bai2025qwen2,
  title={Qwen2. 5-vl technical report},
  author={Bai, Shuai and Chen, Keqin and Liu, Xuejing and Wang, Jialin and Ge, Wenbin and Song, Sibo and Dang, Kai and Wang, Peng and Wang, Shijie and Tang, Jun and others},
  journal={arXiv preprint arXiv:2502.13923},
  year={2025}
}

@article{reimers2019sentence,
  title={Sentence-bert: Sentence embeddings using siamese bert-networks},
  author={Reimers, Nils and Gurevych, Iryna},
  journal={arXiv preprint arXiv:1908.10084},
  year={2019}
}

@misc{flux2024,
    author={Black Forest Labs},
    title={FLUX},
    year={2024},
    howpublished={\url{https://github.com/black-forest-labs/flux}},
}

@article{fu2023dreamsim,
  title={Dreamsim: Learning new dimensions of human visual similarity using synthetic data},
  author={Fu, Stephanie and Tamir, Netanel and Sundaram, Shobhita and Chai, Lucy and Zhang, Richard and Dekel, Tali and Isola, Phillip},
  journal={arXiv preprint arXiv:2306.09344},
  year={2023}
}

@inproceedings{zheng2024llamafactory,
  title={LlamaFactory: Unified Efficient Fine-Tuning of 100+ Language Models},
  author={Yaowei Zheng and Richong Zhang and Junhao Zhang and Yanhan Ye and Zheyan Luo and Zhangchi Feng and Yongqiang Ma},
  booktitle={Proceedings of the 62nd Annual Meeting of the Association for Computational Linguistics (Volume 3: System Demonstrations)},
  address={Bangkok, Thailand},
  publisher={Association for Computational Linguistics},
  year={2024},
  url={http://arxiv.org/abs/2403.13372}
}

@article{anthropic2024claude,
  title={The claude 3 model family: Opus, sonnet, haiku},
  author={Anthropic, AI},
  journal={Claude-3 Model Card},
  volume={1},
  number={1},
  pages={4},
  year={2024}
}

@article{wang2022diffusiondb,
  title={Diffusiondb: A large-scale prompt gallery dataset for text-to-image generative models},
  author={Wang, Zijie J and Montoya, Evan and Munechika, David and Yang, Haoyang and Hoover, Benjamin and Chau, Duen Horng},
  journal={arXiv preprint arXiv:2210.14896},
  year={2022}
}

@inproceedings{lin2014microsoft,
  title={Microsoft coco: Common objects in context},
  author={Lin, Tsung-Yi and Maire, Michael and Belongie, Serge and Hays, James and Perona, Pietro and Ramanan, Deva and Doll{\'a}r, Piotr and Zitnick, C Lawrence},
  booktitle={European conference on computer vision},
  pages={740--755},
  year={2014},
  organization={Springer}
}

@article{mo2025learning,
  title={Learning User Preferences for Image Generation Model},
  author={Mo, Wenyi and Ba, Ying and Zhang, Tianyu and Bai, Yalong and Li, Biye},
  journal={arXiv preprint arXiv:2508.08220},
  year={2025}
}

@article{schuhmann2021laion,
  title={Laion-400m: Open dataset of clip-filtered 400 million image-text pairs},
  author={Schuhmann, Christoph and Vencu, Richard and Beaumont, Romain and Kaczmarczyk, Robert and Mullis, Clayton and Katta, Aarush and Coombes, Theo and Jitsev, Jenia and Komatsuzaki, Aran},
  journal={arXiv preprint arXiv:2111.02114},
  year={2021}
}

@article{hong2025glm,
  title={Glm-4.1 v-thinking: Towards versatile multimodal reasoning with scalable reinforcement learning},
  author={Hong, Wenyi and Yu, Wenmeng and Gu, Xiaotao and Wang, Guo and Gan, Guobing and Tang, Haomiao and Cheng, Jiale and Qi, Ji and Ji, Junhui and Pan, Lihang and others},
  journal={arXiv e-prints},
  pages={arXiv--2507},
  year={2025}
}

@article{wang2025internvl3,
  title={Internvl3. 5: Advancing open-source multimodal models in versatility, reasoning, and efficiency},
  author={Wang, Weiyun and Gao, Zhangwei and Gu, Lixin and Pu, Hengjun and Cui, Long and Wei, Xingguang and Liu, Zhaoyang and Jing, Linglin and Ye, Shenglong and Shao, Jie and others},
  journal={arXiv preprint arXiv:2508.18265},
  year={2025}
}

@inproceedings{chen2025pal,
  title={PAL: Sample-Efficient Personalized Reward Modeling for Pluralistic Alignment},
  author={Chen, Daiwei and Chen, Yi and Rege, Aniket and Wang, Zhi and Vinayak, Ramya Korlakai},
  booktitle={The Thirteenth International Conference on Learning Representations},
  year={2025}
}

@article{mo2025prefgen,
  title={PrefGen: Multimodal Preference Learning for Preference-Conditioned Image Generation},
  author={Mo, Wenyi and Zhang, Tianyu and Bai, Yalong and Han, Ligong and Ba, Ying and Metaxas, Dimitris N},
  journal={arXiv preprint arXiv:2512.06020},
  year={2025}
}
